\documentclass[10pt,journal,compsoc]{IEEEtran}

\usepackage{ifpdf}
\ifpdf
  \pdfoutput=1\relax                   
  \usepackage{graphicx}                
  \DeclareGraphicsExtensions{.pdf,.png,.jpg,.jpeg} 
\else
  \usepackage{graphicx}                
  \DeclareGraphicsExtensions{.eps}     
\fi%

\graphicspath{{figures/}{pictures/}{images/}{./}} 

\usepackage{microtype}                 
\PassOptionsToPackage{warn}{textcomp}  
\usepackage{textcomp}                  
\usepackage{mathptmx}                  
\usepackage{times}                     
\usepackage{cite}                      
\usepackage{tabu}                      
\usepackage{booktabs}                  
\usepackage{multirow}
\usepackage{amsfonts}
\usepackage{newtxmath}
\usepackage{bigdelim}
\usepackage{bm}
\usepackage{wrapfig}

\usepackage{color}

\begin{document}
%
\title{Laplacian2Mesh: Laplacian-Based Mesh Understanding}
%
%
\author{Qiujie~Dong,
Zixiong~Wang,
Manyi~Li,
Junjie~Gao,
Shuangmin~Chen,
Zhenyu~Shu,
Shiqing~Xin$^\star$,
Changhe~Tu,
Wenping~Wang,~\IEEEmembership{Fellow,~IEEE}

\IEEEcompsocitemizethanks{
\IEEEcompsocthanksitem Q. Dong, Z. Wang, J. Gao, S. Xin, and C. Tu are with the School of Computer Science and Technology, Shandong University, Qingdao, Shandong, China.
E-mail: qiujie.jay.dong@gmail.com, zixiong$\_$wang@outlook.com, gjjsdnu@163.com, xinshiqing@sdu.edu.cn, chtu@sdu.edu.cn.

\IEEEcompsocthanksitem M. Li is with the School of Software, Shandong University, Jinan, Shandong, China. E-mail: manyili@sdu.edu.cn.

\IEEEcompsocthanksitem S. Chen is with the School of Information and Technology, Qingdao University of Science and Technology, Qingdao, Shandong, China. E-mail: csmqq@163.com.

\IEEEcompsocthanksitem  Z. Shu is with the School of Computer and Data Engineering, Ningbo Institute of Technology, Zhejiang University, Ningbo, Zhejiang, China. E-mail: shuzhenyu@nit.zju.edu.cn.

\IEEEcompsocthanksitem  W. Wang is with the Department of Computer Science $\&$ Engineering, Texas A$\&$M University. E-mail: wenping@cs.hku.hk.

\IEEEcompsocthanksitem
S. Xin is the corresponding author.
}
\thanks{Manuscript received April 19, 2005; revised August 26, 2015.}}

\markboth{Journal of \LaTeX\ Class Files,~Vol.~14, No.~8, August~2015}%
{Dong \MakeLowercase{\textit{et al.}}: Laplacian2Mesh: Laplacian-Based Mesh Understanding}

\IEEEtitleabstractindextext{%
\begin{abstract}
Geometric deep learning has sparked a rising interest in computer graphics to perform shape understanding tasks, such as shape classification and semantic segmentation.
When the input is a polygonal surface,
one has to suffer from the irregular mesh structure. Motivated by the geometric spectral theory, we introduce {\em Laplacian2Mesh}, a novel and flexible convolutional neural network (CNN) framework for coping with irregular triangle meshes (vertices may have any valence).
By mapping the input mesh surface to the multi-dimensional Laplacian-Beltrami space, Laplacian2Mesh enables one to perform shape analysis tasks directly using the mature CNNs, without the need to deal with the irregular connectivity of the mesh structure. 
We further define a mesh pooling operation such that the receptive field of the network can be expanded while retaining the original vertex set as well as the connections between them. Besides, we introduce a channel-wise self-attention block to learn the individual importance of feature ingredients. Laplacian2Mesh not only decouples the geometry from the irregular connectivity of the mesh structure but also better captures the global features that are central to shape classification and segmentation. Extensive tests on various datasets demonstrate the effectiveness and efficiency of Laplacian2Mesh, particularly in terms of the capability of being vulnerable to noise to fulfill various learning tasks.
\end{abstract}

}

\maketitle

%
\IEEEpeerreviewmaketitle

\IEEEraisesectionheading{
\section{Introduction}
\label{sec:introduction}}

\IEEEPARstart{T}he rapidly emerging 3D geometric learning techniques 
have achieved impressive performance in various applications, such as classification~\cite{Su2015MultiviewCN, Wu20153DSA}, semantic segmentation~\cite{Hanocka2019MeshCNNAN, Tchapmi2017SEGCloudSS}, and shape reconstruction~\cite{Brock2016GenerativeAD, Liu2021DeepIM}. 
A large number of deep neural networks have been designed to deal with 3D shapes of various representations, including voxels~\cite{Klokov2017EscapeFC, Wang-2017-ocnn}, multi-view images~\cite{Le2017AMR, Su2015MultiviewCN}, point clouds~\cite{Qi2017PointNetDL, Qi2017PointNetDH}, meshes~\cite{Hanocka2019MeshCNNAN, Lahav2020MeshWalkerDM}, etc. 
Among all the representations, 
polygonal surfaces, as one of the most popular shape representations, 
are flexible in characterizing an arbitrarily complex 3D shape in an unambiguous manner.
However, the irregular connections between vertices make designing learning-based methods extremely difficult.
To be more specific, the challenges include the following aspects: 
First, polygonal surfaces that represent the same shape may have varying numbers of vertices and faces.
Second, the valence of a vertex is not fixed, making it hard to define a regular structure as the receptive field for the conventional convolution operator.
Finally, even if the number of vertices is given,
various triangulation forms may exist while the difference in representation accuracy 
is neglectable.

In the research community, there have been many attempts for developing mesh-specific convolution and pooling/unpooling operations. For example, GeodesicCNN~\cite{Masci2015GeodesicCN} and CurvaNet~\cite{He2020CurvaNetGD} 
are proposed to define convolution kernels on the parameterization plane for each 
local surface patch.
MeshCNN~\cite{Hanocka2019MeshCNNAN} and PD-MeshNet~\cite{Milano2020PrimalDualMC} 
take a mesh as a graph with regard to the vertex set.
They introduce edge-contraction-based pooling/unpooling operations to deal with the 
mesh-based deep learning tasks.
Very recently, SubdivNet~\cite{Hu2022SubdivNet}
suggests a convolutional operation by referring to the coarse-to-fine Loop subdivision algorithm.
All the above-mentioned approaches have to invent an artificial convolutional scheme and explicitly come into contact with
the irregular triangulation, imposing harsh requirements on the input mesh and suffering from various mesh imperfections. 

In this paper, we propose {\em Laplacian2Mesh},
inheriting the spirit of those spectral approaches,
to make it possible to conduct learning tasks in the spectral domain
when the input is a polygonal mesh. 
Note that
the main technique of spectral approaches 
is to encode the overall shape by a subset of the eigenvectors decomposed from the Laplacian matrix of the input mesh. 
Two reasons account for why we advocate using the new representation in deep learning.
First, as the low-frequency signals, given by those eigenvectors with small eigenvalues, encode the overall shape, they are more semantically important than high-frequency signals for most understanding tasks. 
It is easy for one to separate low-frequency signals from high-frequency signals in spectral approaches by setting a simple parameter $k$. 
Second, the Laplacian-based spectral transform can decouple shape variations from the tedious triangulation, avoiding tangling with tedious and irregular triangulations. 
Even if the input mesh contains a limited number of defects (e.g., non-manifold),
the representation still works.

Some researchers have turned their attention to this promising research area. 
For example, the recent DiffusionNet~\cite{Nicholas2022DiffusionNet} took the vertex positions and the spatial gradients as the original signal and then projected them w.r.t.~the Laplacian-based spectral basis. It is pointed out in this work that directly learning the frequency signals suffers from the different eigenvectors decomposed from different shapes. They merely use the spectral transform to facilitate
the spectral acceleration and project the spectral features back to the spatial domain, yielding the vertex-wise features,which means that it is not
a spectral learning method but just a numerical scheme to compute diffusion.

Therefore, there is still a long way to go to develop a deep network for the spectral signals of the polygonal meshes.
The first difficulty lies in that different shapes have different eigenvectors~\cite{Hanocka2021AnIT, Nicholas2022DiffusionNet}, resulting in a learned spectral convolution filter on one shape that cannot generalize to a different shape. 
The second difficulty is how to bring this kind of network to its full potential,
i.e., learning the useful features for understanding a shape. 
Potentially helpful network architecture is the famous U-Net~\cite{Ronneberger2015UNetCN} with the encoder part and the decoder part,
where the encoder consists of the convolutional layers followed by pooling operation,
and the decoder, also consisting of multiple connected layers, uses transposed convolution to permit localization. 

Our Laplacian2Mesh is designed to deal with the above-mentioned two difficulties by transforming the features of the polygonal meshes into eigenspace~\cite{Lvy2010SpectralMP}. 
On the one hand, we select three groups of eigenvectors to construct the multi-resolution bases. We use the Squeeze-and-Excitation ResNet (SE-ResNet)~\cite{Hu2020SqueezeandExcitationN} as the basic unit to 
define the feature extractor, as well as the carefully designed Laplacian pooling/unpooling operations to fuse the resulting features with different dimensions.
Additionally, we consider
the shape descriptors (vertex normals, dihedral angles of vertices, Gaussian curvature, Laplacian eigenvectors, heat kernel signature (HKS)~\cite{Sun2009ACA}) and the pose descriptor (vertex coordinates) at the same time to feed the network so that useful shape information can be fully utilized. 

We conduct comprehensive ablation/comparison experiments to validate the effectiveness of our approach. 
Tests on various shape understanding tasks, including shape classification and semantic segmentation, show that our algorithm outperforms the state-of-the-art on average. The biggest advantage of our approach lies in the robustness to noise. Besides, our approach can tolerate a limited number of mesh defects. 

\begin{figure*}[ht]
 \begin{center}
 \includegraphics[width=\linewidth]{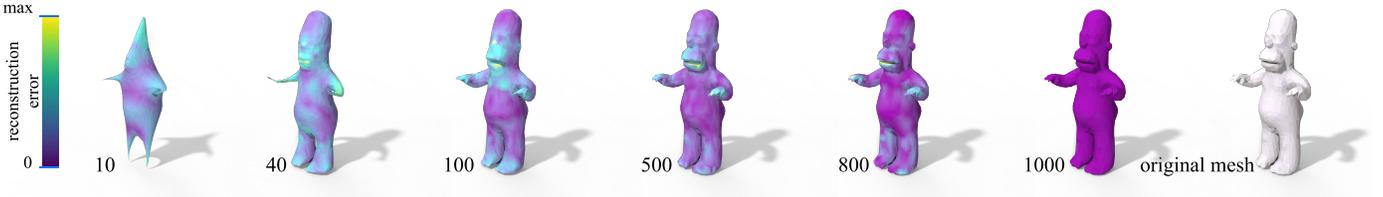}
 \end{center}
 \caption{The spectral surface reconstruction when selecting different numbers of Laplacian eigenvectors as the spectral basis. From left to the right, the fewer smallest eigenvectors used, the less high-frequency features preserved in the reconstruction. Therefore, by selecting the proper basis, the spectral transforms not only encode the irregular mesh structure into fix-sized signals but also are resistant to the high-frequency noise for the mesh understanding tasks.
 }
 \label{fig:reconstruct_homer}
\end{figure*}

\section{Related work}
\label{sec:related_work}

In this section, we first briefly introduce the related works on the 3D geometric learning~\cite{Bronstein2021GeometricDL} with various shape representations. We refer the readers to~\cite{Bronstein2017GeometricDL, Xiao2020ASO} for a more comprehensive survey. After that, we review the learning-based mesh understanding methods, which are highly relevant to the theme of this paper. Finally, we talk about
the spectral-based technique and its applications.

\subsection{3D Geometric Deep Learning}
The design of geometric deep learning networks relies heavily on the shape representations. Wu et al.~\cite{Wu20153DSA} proposed a 3D convolutional network based on the voxel representation for the shape classification and retrieval tasks. Due to their regular data structure, 3D voxels are easily processed by different 3D convolutional networks, making the voxel representation  frequently utilized in the applications such as shape reconstruction~\cite{Brock2016GenerativeAD}, semantic segmentation~\cite{Tchapmi2017SEGCloudSS}, and alignment~\cite{Hanocka2019ALIGNetPA}.
Its usage is greatly limited by the cubically increasing complexity in the memory and computation cost. To deal with this issue, the octree-based volumetric representation~\cite{Riegler2017OctNetLD, Wang2017OCNN} 
is more compact and has received more attention recently. Through parameterization, a surface can be encoded by an image-like data structure,
e.g. the geometry images~\cite{Sinha2016DeepL3} and the toric covers~\cite{Haim2019SurfaceNV, Maron2017ConvolutionalNN}.
Another way is to render the 3D shapes as multi-view images,
so that 2D CNN can be directly used for the understanding tasks~\cite{Le2017AMR, Su2015MultiviewCN}. However, 
it may fail for a complicated shape with a highly curved surface.

Point cloud is also a popular representation in 3D geometric learning.
For example, PointNet~\cite{Qi2017PointNetDL} and PointNet++~\cite{Qi2017PointNetDH} use the KNN search to build the spatial correlations between points for feature extraction.
Some following works, such as PointCNN~\cite{Li2018PointCNNCO} and KPConv~\cite{Thomas2019KPConvFA}, further improve feature learning from point cloud.
Recently, PCT~\cite{Guo2021PCTPC} applies the attention-based transformer to deal with point cloud, achieving state-of-the-art performance. 
However, for a point cloud, the direct exploitation of the geometry is challenging due to inherent obstacles such as noise, occlusions, sparsity or variance in the density. 
On the other hand, polygonal surfaces typically offer more detailed geometry and topology descriptions, hopefully improving the understanding of behavior.

\subsection{Learning-based Mesh Understanding Methods}

The mesh representation is often considered as a graph structure. MeshWalker~\cite{Lahav2020MeshWalkerDM} explores the mesh geometry and topology via random walks. The Recurrent Neural Network (RNN) is used to learn the deep features along each walk. Some other methods construct the local relationship between the vertices by considering the K-ring neighborhood. Masci et al.~\cite{Masci2015GeodesicCN} proposed to flatten each of the patches with small curvature to the 2D plane, and apply the geodesic convolutional neural network on the non-Euclidean manifolds. He et al.~\cite{He2020CurvaNetGD} defined a directional curvature along with rotation in the tangent plane to obtain neighbor information with a consistent structure. These methods focus on defining the localized convolution kernels.
Some similar strategies include local spectral filters~\cite{Boscaini2015LearningCD} and local geometric descriptors (e.g. B-splines~\cite{Fey2018SplineCNNFG}, wavelets~\cite{Schonsheck2018ParallelTC} and extrinsic Euclidean convolution~\cite{Schult2020DualConvMeshNetJG}) to align a local patch with the convolution kernel.

Some recent works leverage the classical mesh processing techniques to 
define the key modules of a geometric deep learning network, such as the convolutional operation and the pooling operation. 
For example, MeshCNN~\cite{Hanocka2019MeshCNNAN} defines the receptive field for the convolutional operation, based on the observation that each edge is correlated with two neighbor triangular faces. Furthermore, MeshCNN defines the unique pooling operation based on the edge-collapse algorithm.
The PD-MeshNet~\cite{Milano2020PrimalDualMC} constructs the primal graph and the dual graph to capture the adjacency information encoded in the triangle mesh.
The pooling operation of PD-MeshNet is implemented based on mesh simplification. 
Benefiting from the up-sampling process of the Loop subdivision algorithm, SubdivNet~\cite{Hu2022SubdivNet} proposes a mesh convolution operator that allows one to aggregate local features from nearby faces. But it requires the input to hold the subdivision connectivity.
All the aforementioned methods focus on the spatial structure
and have to handle the irregular connections.

A very recent work DiffusionNet~\cite{Nicholas2022DiffusionNet} uses the spectral acceleration technique to diffuse the vertex features, showing a great potential of the spectral analysis in mesh-based learning tasks. 
We propose to extend the backbone from the spatial to the spectral domain, and further improve the performance of shape understanding, particularly in the presence of noise.

\subsection{Spectral Surface Representation}
As pointed out in~\cite{Bronstein2017GeometricDL},
spectral surface representation is important in encoding the shape information with varying frequencies, regardless of the connection structure. 
In the discrete setting, one can take the eigenvectors of the Laplacian matrix 
as the basis for the spectral transform. 
As shown in Figure~\ref{fig:reconstruct_homer}, the low-frequency signals, given by the eigenvectors with the $k$ smallest eigenvalues, can capture the essential shape information,
as long as $k$ is large enough. 
Therefore, by selecting the proper number of eigenvectors, the spectral projection can be seen as a low-pass filter to the mesh feature signals, naturally resisting noise and sampling bias. 
The spectral analysis plays a central role in many geometry processing and shape analysis tasks, such as shape segmentation~\cite{Katz2003HierarchicalMD}, symmetry detection~\cite{Ovsjanikov2008GlobalIS}, shape correspondence~\cite{FunctionalMaps2012}, shape recognition~\cite{Bronstein2011ShapeRW}, and shape retrieval~\cite{Bronstein2011ShapeGG}.

Some existing research works~\cite{Lvy2010SpectralMP, Karni2000SpectralCO}
show that it is possible to use the spectral analysis technique to define a compact representation of a polygonal surface. 
Heat Kernel Signature (HKS)~\cite{Sun2009ACA} inspires us in that
the local-to-global shape variations can be encoded through heat diffusion.
Please refer to~\cite{Wang2019IntrinsicAE,Lvy2010SpectralMP} 
for a more detailed survey.

There are already some works using spectral signals in deep learning. The spectral-based graph networks, e.g., SpectralCNN\cite{estrach2014spectral} and ChebNet~\cite{defferrard2016convolutional}, convert the graph structure into the spectral domain and perform the multiplication on the spectral signals. 
LaplacianNet~\cite{Qiao2022LearningO3} uses the graph Laplacian-based spectral clustering for over-segmentation, followed by performing a max-pooling operation on each class to obtain local information. LaplacianNet includes the Correlation Net as a component to learn a correlation matrix to fuse features across clusters.
But it's worth noting that LaplacianNet still focuses on the spatial relation among the local patches.

\section{Methodology}
\label{sec:methodology}

We compute a collection of intrinsic and extrinsic mesh features, and transform them into the spectral domain to feed the network. Then the network
utilizes the U-Net architecture to handle various mesh understanding tasks. 
Additionally, we develop a channel-wise attention mechanism employing the small-sized Squeeze-and-Excitation blocks (SE-block)~\cite{Hu2020SqueezeandExcitationN}, before fusing features of various resolutions using Laplacian pooling/unpooling. 
The whole network is connected by different head blocks.
Finally, different loss functions are proposed to cope with mesh classification and segmentation tasks, respectively.

\begin{figure}[tb]
 \begin{center}
 \includegraphics[width=\columnwidth]{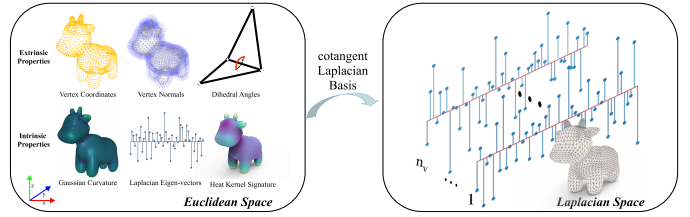}
 \end{center}
 \caption{
 We compute a set of extrinsic and intrinsic geometric features on the mesh surface and convert them into spectral signals as the input of our network. 
 }
 \label{fig:laplacian_space}
\end{figure}

\begin{figure*}[tb]
\begin{center}
 \includegraphics[width=\linewidth]{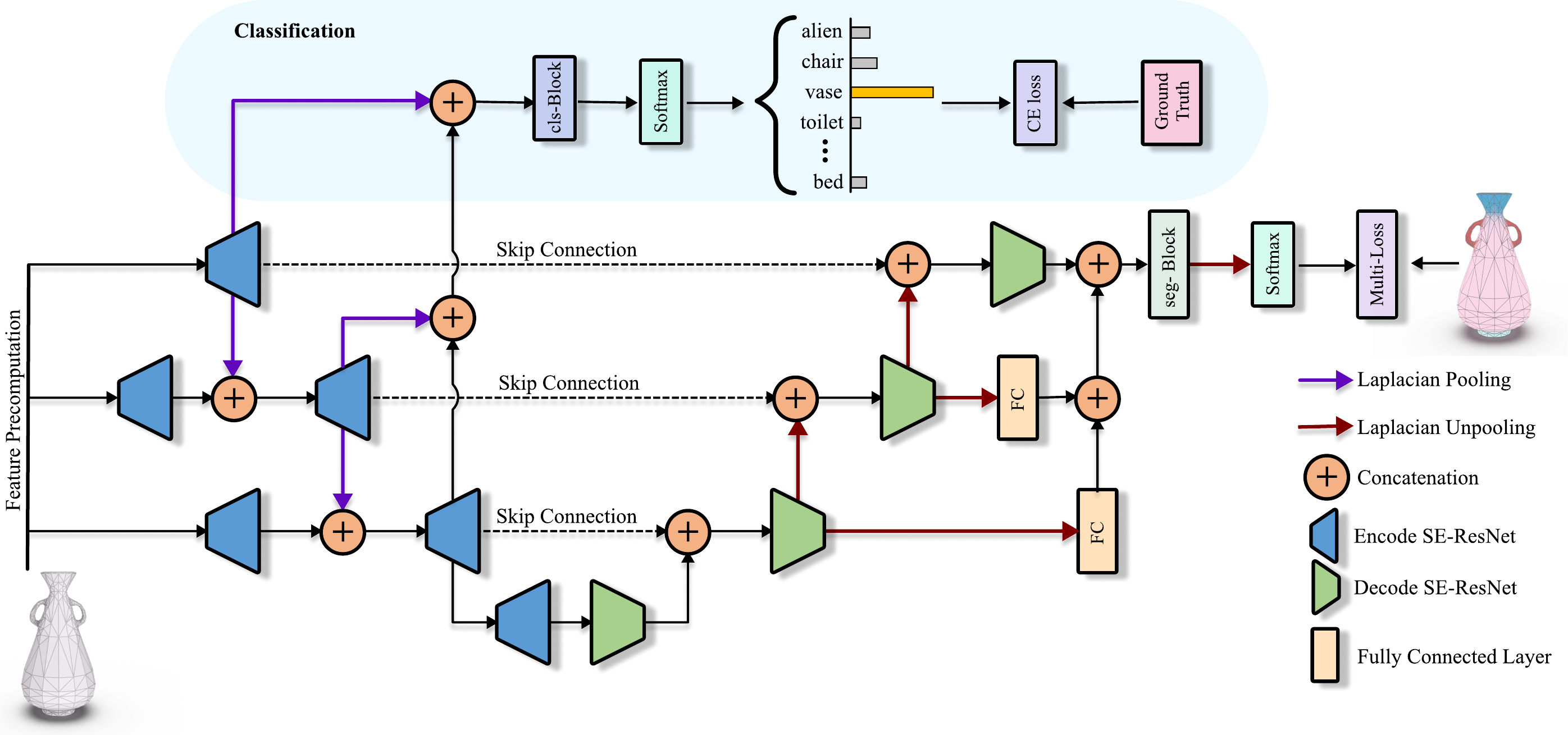}
\end{center}
  \caption{
  Our network pipeline for coping with the mesh classification and segmentation tasks. Given a 3D mesh as the input, we precompute the extrinsic and intrinsic geometric features and project them into the spectral domain w.r.t. three different resolutions. Inspired by the U-Net architecture, we propose to use the SE-ResNet blocks with small-sized convolution kernels to fuse the nearby-frequency features, and the Laplacian pooling/unpooling to fuse the spectral features of different resolutions. For the segmentation task, we re-scale (with the yellow block) and concatenate the features together to be processed by the segmentation block. 
  }
\label{fig:pipeline}
\end{figure*}

\subsection{Laplacian spectral transform}
\label{sec:spectral_transform}

Given a triangle mesh with $n$ vertices, the Laplacian matrix~\cite{Meyer2002DiscreteDO, Rustamov2007LaplaceBeltramiEF} can be written as:
\begin{equation}
\label{equ:vertex_example}
\mathbf{L} = \mathbf{M}^{-1} \mathbf{C},
\end{equation}
where $\mathbf{M} \in \mathbb{R}^{n \times n}$ is the diagonal matrix whose $i$-th entry along the diagonal is twice the influence area of the vertex $v_i$, $\mathbf{C}$ is the sparse cotangent 
weighted  matrix as in Eq.~(\ref{equ:cotangent_matrix}):
\begin{equation}
\label{equ:cotangent_matrix}
\mathbf{C}_{ij} =
\begin{cases}
- (\cot \alpha_{ij} + \cot  \beta_{ij}), &  i\neq j, v_j \in r_1(v_i) \\
\sum_{v_j \in r_1(v_i)} (\cot \alpha_{ij} + \cot \beta_{ij}), &  i = j \\
0, & v_j \notin r_1(v_i),
\end{cases}
\end{equation}
where $r_1(v_i)$ is the set of all the 1-ring adjacent vertices of $v_i$, 
and $\alpha_{ij}$ and $\beta_{ij}$ are the two angles opposite to edge $v_iv_j$. Please refer to our Appendices for more details.

The Laplacian matrix completely encodes the intrinsic geometry.
The eigendecomposition of the Laplacian matrix $\mathbf{L}$ enables the transformation between the spatial and the spectral domain. Specifically, after performing the eigendecomposition, we can sort and select the k eigenvectors $\mathbf{\Phi}_k=[\bm{\phi}_0, \bm{\phi}_1, \dots, \bm{\phi}_{k-1}]$ corresponding to the $k$ smallest eigenvalues.
It's worth noting that the smallest eigenvalue is~0.
$\mathbf{\Phi}_k$ can be understood as a low-frequency filter. 
Suppose that we define a scalar field~$f$ on the surface.
It can be decomposed into a combination of the eigenvalues.
If we keep only the part spanned by $\mathbf{\Phi}_k$,
we obtain the reconstructed counterpart of~$f$.

Figure~\ref{fig:reconstruct_homer} shows the Laplacian spectral reconstruction with different numbers of eigenvectors as the basis. The mesh vertex set $\mathbf{V}$ is first projected to $\widetilde{\mathbf{V}}$ in the spectral domain, and then reconstructed as $\widehat{\mathbf{V}}$ following Eq.~(\ref{equ:vertex_example}):
\begin{equation}
\label{equ:vertex_example}
\begin{split}
&\widetilde{\mathbf{V}} = \mathbf{\Phi}_{k}^T \mathbf{V}, \\
&\widehat{\mathbf{V}}=\mathbf{\Phi}_{k} (\mathbf{\Phi}_{k}^T \mathbf{V}).
\end{split}
\end{equation}
Figure~\ref{fig:reconstruct_homer} also illustrates that the spectral transform can
serve as a low-pass filter of the mesh feature signal. Therefore, it prevents the mesh processing from being affected by the high-frequency noise.

We take $\mathbf{\Phi}_{k}$ as the basis to perform the Laplacian spectral transform. To simplify the computation, the eigenvector matrix $\mathbf{\Phi}$ is obtained by performing the eigendecomposition on the cotangent weight matrix instead of the discrete Laplace-Beltrami matrix itself. 
Please refer to Appendices for more detailed explanation.

\begin{figure}[tb]
 \begin{center}
 \includegraphics[width=\columnwidth]{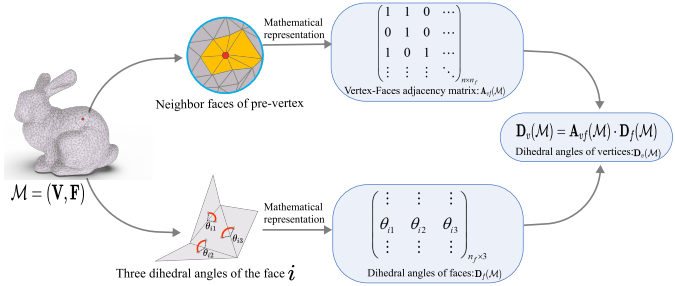}
 \end{center}
 \caption{
 The vertex-wise feature based on the dihedral angles, which are originally defined on the shared edges between faces. We use the vertex-face adjacency matrix to average the three dihedral angles of the one-ring faces to form the vertex-wise dihedral angle.
 }
 \label{fig:dihedral_angle_vertices}
\end{figure}

\subsection{Input Features}
\label{sec:input_features}

In geometric deep learning, 
a commonly used way is to extract various geometric features to define a synthesized
shape descriptor. The intrinsic and extrinsic features characterize 
the shape from different perspectives, and their combination can
describe the shape~\cite{Hanocka2019MeshCNNAN, Hu2022SubdivNet, Qiao2022LearningO3, Nicholas2022DiffusionNet}. 

Our input feature $\mathbf{G} \in \mathbb{R}^{n\times39}$ in the spatial domain is composed of the per-vertex features, each of which is a concatenation of a 30-dimensional intrinsic shape descriptor and a 9-dimensional extrinsic shape descriptor (see Figure~\ref{fig:laplacian_space}). The intrinsic shape descriptor is formed by the 1-dimensional Gaussian curvature, the 9-dimensional HKS (Heat Kernel Signature)~\cite{Sun2009ACA}, and 
low-frequency eigenvectors of the cotangent weight matrix corresponding to the 20 lowest frequencies excluding 0. The extrinsic shape descriptor includes the 3-dimensional vertex coordinates, the 3-dimensional vertex normal, and the 3-dimensional dihedral angles for each vertex (we will explain it later).
Finally, the input feature vector $\mathbf{G}$,
upon being mapped to the Laplacian spectral domain, becomes:
\begin{equation}
\label{equ:Laplacian_feature}
\widetilde{\mathbf{G}} = \mathbf{\Phi}_{k}^T \mathbf{G}.
\end{equation}
To this end, an arbitrary polygonal surface 
is transformed into a $k\times 39 $ matrix~$\widetilde{\mathbf{G}}$, regardless of the geometry, topology, connection structure, and mesh complexity.

It is worth noting that the dihedral angles are originally defined as the angle between  two adjacent faces sharing a common edge. We extend the concept to a per-vertex feature by distributing the dihedral angle of a mesh edge to the endpoints 
and the opposite vertices. 
Recall that the face~$f=\triangle v_1v_2v_3$ has three bounding edges~$\overrightarrow{v_2v_3},\overrightarrow{v_3v_1},\overrightarrow{v_1v_2}$, and they give three dihedral angles~$\theta_1,\theta_2,\theta_3$, respectively.
In our scenario, we need to define vertex-wise geometric features.
For this purpose, we can relate the dihedral angles to the three vertices of~$f$.
Particularly, we assign~$\theta_1,\theta_2,\theta_3$ to~$v_1$,
assign~$\theta_2,\theta_3,\theta_1$ to~$v_2$,
and assign~$\theta_3,\theta_1,\theta_2$ to~$v_3$.
This can be implemented by a simple matrix multiplication.
As illustrated in Figure~\ref{fig:dihedral_angle_vertices}, the vertex-based dihedral-angle
matrix~$\mathbf{D}_v(\mathcal{M})$ is defined as the multiplication of the vertex-face adjacency matrix $\mathbf{A}_{vf}(\mathcal{M})$ and the face-based dihedral-angle matrix~$\mathbf{D}_f(\mathcal{M})$, i.e.,
\begin{equation}
\label{equ:dihedral_angles_vertices}
\mathbf{D}_v(\mathcal{M}) = \mathbf{A}_{vf}(\mathcal{M}) \cdot \mathbf{D}_f(\mathcal{M}).
\end{equation}

\subsection{Network}
\label{subsec:overview}
We feed the aforementioned feature matrix $\widetilde{\mathbf{G}}$ 
into the network in a multi-resolution manner. Specifically, we select three different resolutions of $\widetilde{\mathbf{G}}$ in a decreasing order,
where the dominant dimensions of $\widetilde{\mathbf{G}}$ are denoted as $\mathbf{k}=\{k_i \mid i=0,1,2\}$. We then select the  spectral basis $\mathbf{\Phi}_{k_i}$ and compute the corresponding feature matrix $\widetilde{\mathbf{G}}$, denoted as $\widetilde{\mathbf{G}}_0$, $\widetilde{\mathbf{G}}_1$, $\widetilde{\mathbf{G}}_2$, respectively. 
Our empirical evidence shows that $k_0$, i.e., the largest one, should be greater than or equal to $n_v / 2$~\cite{Lvy2010SpectralMP}, where $n_v$ is the maximum total number of vertices for the 3D models in the dataset.

As shown in Figure~\ref{fig:pipeline}, our network resembles the U-Net structure. The three spectral feature matrices, $\widetilde{\mathbf{G}}_0$, $\widetilde{\mathbf{G}}_1$, $\widetilde{\mathbf{G}}_2$, are processed by the corresponding convolutional blocks independently and then fused from the higher resolution to the lower resolution. We use the mirror structure during decoding, while taking the encoding feature at the same resolution via the skip connection at each level. The network can be easily instantiated for different tasks by adding the head blocks; See the classification block or the segmentation block in Figure~\ref{fig:pipeline}. The main difference between ours and the existing U-Net networks lies in the SE-blocks and the proposed Laplacian pooling/unpooling operations, as described later.

\begin{figure}[tb]
 \centering
 \includegraphics[width=\columnwidth]{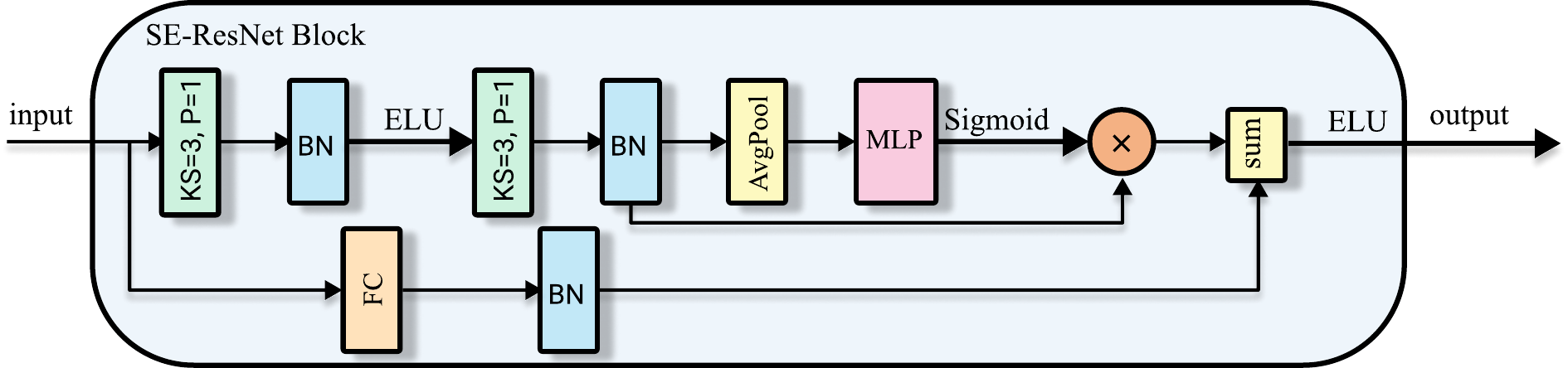}
 \caption{The layers of SE-ResNet Block.}
 \label{fig:se_resnet}
\end{figure}

\subsubsection{SE-ResNet Block}
Instead of learning the multiplication in the spectral domain as some spectral-based networks~\cite{Kipf2017SemiSupervisedCW, Litman2014LearningSD} do, we apply the convolution operations to the spectral signals. However, due to the lack of shift-invariance in the spectral domain, the commonly-used convolutions with large receptive fields for pattern recognition are not suitable. 
Here, we use the convolution with small kernels. 
The small-sized convolution kernel acts as an aggregation function to fuse the close frequency features.

In addition, we take the Squeeze-and-Excitation ResNet (SE-ResNet)~\cite{Hu2020SqueezeandExcitationN} (see Figure~\ref{fig:se_resnet}) as the basic unit to process the spectral signals.
The Squeeze-and-Excitation (SE) module first performs the squeeze operation on the feature map obtained by convolution to obtain the channel-level global features. In order to obtain the weights of different channels, the excitation operation on the global features is then performed to learn the relationship between different channels. Finally, the channel weights are multiplied by the original feature map to get the final features. In its essence, the SE module performs the attention or gating operations on the channel dimension. This attention mechanism allows the model to pay more attention to the significant channel features while suppressing those less important channel features. The module alleviates the tedious work of manual feature selection and improves the representational capacity of a network by performing dynamic channel-wise feature recalibration.

Unlike the other networks that directly apply the same trainable network weights on the feature map of various data samples, the SE-block predicts the adaptive weights to fuse the channel-wise features, and the residual connection helps avoid the vanishing gradients of the deep network. We experimentally found that the SE-block achieves the best performance with the kernel size $ks=3$ and padding $p=1$ while the layers without the SE-block achieve the best performance when $ks=1$ and $p=0$.

\begin{figure}[t]                                                        
 \begin{center}
 \includegraphics[width=\columnwidth]{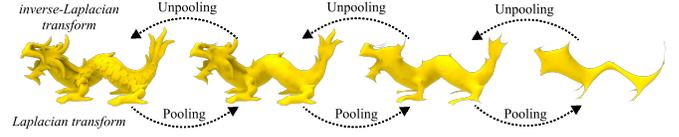}
 \end{center}
 \caption{
 The Laplacian pooling and unpooling operations 
 transform the spectral-based features between different resolutions,
 where the pooling operation proceeds from a finer level to a coarser level
 while the unpooling operation does the opposite.
 }
 \label{fig:pooling_unpooling}
\end{figure}

\subsubsection{Laplacian Pooling and Unpooling}

Note that the input of our network is the multi-resolution spectral signals with different sizes of eigenvector basis, which is obtained by Eq.~(\ref{equ:vertex_example}). 
The Laplacian pooling and unpooling operations are necessary 
to fuse them together.
Figure~\ref{fig:pooling_unpooling} illustrates how the Laplacian pooling and unpooling operations transform the signals between the spectral basis with different resolutions.

For the $i$-th layer and the $j$-th layer in the network,
we propagate the output of the $i$-th layer 
to the $j$-th layer based on $\mathbf{\Phi}_{k_i}$ and $\mathbf{\Phi}_{k_j}$:
\begin{equation}
\mathbf{\Gamma}_{ij} = (\mathbf{\Phi}_{k_j}^T \mathbf{\Phi}_{k_i}) \mathbf{\Gamma}_i.
\end{equation}
The transformed feature $\mathbf{\Gamma}_{ij}$ is then concatenated with the $j$-th layer, yielding a concatenated feature:
\begin{equation}
\widetilde{\mathbf{\Gamma}}_{j} = \mathbf{\Gamma}_{j} \oplus \mathbf{\Gamma}_{ij},
\end{equation}
which shows the mechanism of our Laplacian pooling operation when the resolution of $i$-th layer is larger, and Laplacian unpooling operation otherwise.

\begin{figure}[b]
 \centering
 \includegraphics[width=\columnwidth]{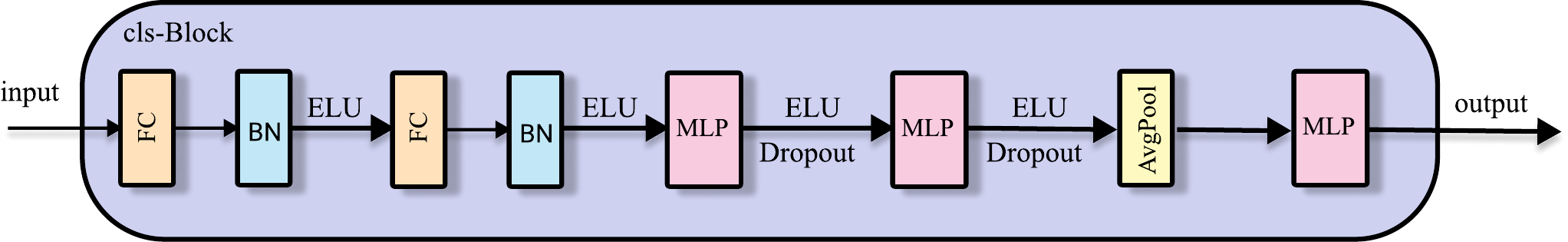}
 \caption{The head block for the classification task.}
 \label{fig:cls-Block}
\end{figure}

\begin{figure}[b]
 \centering
 \includegraphics[width=\columnwidth]{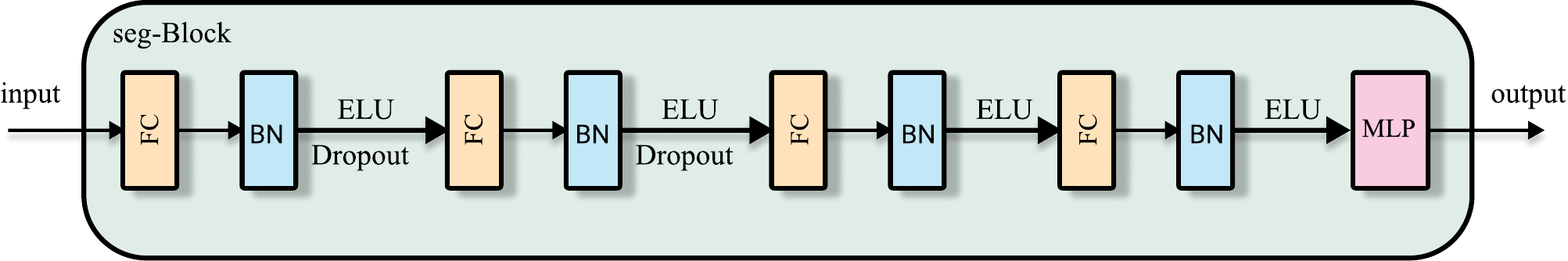}
 \caption{The head block for the semantic segmentation task.}
 \label{fig:seg-block}
\end{figure}

\subsubsection{Classification and Segmentation Modules}
It's natural that our Laplacian2Mesh network can deal with different mesh understanding tasks if the head blocks are carefully set.
Figure~\ref{fig:pipeline} illustrates the classification and segmentation blocks, as well as their connection with the backbone network. For the classification task, we only make use of the encoding stage of our network and fuse the multi-resolution features to form the input of the classification block (cls-block). The classification block follows the popular implementation in the related works~\cite{Milano2020PrimalDualMC}, and uses the simple Global Average Pooling and softmax equipped MLP to obtain the final prediction. As for the segmentation task, it has to utilize the entire U-Net architecture and 
then add a segmentation block (seg-block) to compute the segmentation-aware features in the spectral domain. The spectral feature is then transformed to the spatial domain.
Finally, we use a softmax function to obtain the segmentation label for each vertex.

We carefully designed cls-Block (see Figure~\ref{fig:cls-Block}) and seg-Block (see Figure~\ref{fig:seg-block}) for the tasks of shape classification and semantic segmentation, respectively. In these blocks, $ELU(\cdot)$~\cite{Clevert2016FastAA} is chosen as the activation function in our Laplacian2Mesh, which eliminates the influence of bias shift and improves the robustness to noise.

\subsection{Loss Function}
\label{subsec:loss_function}
For the shape classification task,
one can simply use the cross-entropy loss between the predicted label and the ground-truth.
For the shape segmentation task,
however, we need to add one additional loss term to ensure 
the spatial coherence of the vertex-based part labels.
Therefore, our loss function for the mesh segmentation task can be written as
\begin{equation}
\mathcal{L} = \mu \cdot \mathcal{L}_{\text{ce}} + \nu \cdot \mathcal{L}_{\text{adj}},
\end{equation}
where $\mathcal{L}_{\text{ce}}$ is the segmentation loss to minimize the error between the predicted vertex label and the ground-truth, $\mathcal{L}_{\text{adj}}$ is the adjacency loss to encourage the label coherence between neighboring vertices, and $\mu,\nu$ are a pair of hyperparameters to balance the two terms.

\textbf{Segmentation loss.} 
Note that the output of the segmentation block 
still encodes the segmentation information in the spectral domain. 
We need to transform it to the spatial domain with Eq.~(\ref{equ:vertex_example}).
For each vertex, we obtain a one-hot vector to characterize 
the probability in which the vertex belongs to each part,
and assign the vertex to the part with the maximum probability. 
The segmentation loss is then defined as the cross-entropy loss between the predicted labels and the ground-truth labels.

\textbf{Adjacency loss.} 
Intuitively speaking, 
a pair of neighboring vertices tend to own consistent labels.
Therefore, it is necessary to enforce the label coherence in the one-ring neighborhood for each vertex. 
We introduce three matrices, i.e., $\mathbf{\Psi}, \mathbf{\Omega}$ and $\mathbf{A}$, to weigh the influence between neighboring vertices.
Let $\mathbf{\Psi}$ be the all-pair straight-line distance matrix between vertices,
satisfying $\mathbf{\Psi}=\left\{ \|\mathbf{v}_i - \mathbf{v}_j \| \mid i,j\in\{0,1,\cdots, n-1\} \right\}$.
Let $\mathbf{\Omega}=\exp(-\mathbf{\Psi} / (2\delta))$ be the Gaussian filter with a bandwidth $\delta$.
Let $\mathbf{A}$ be the adjacency matrix of the mesh. 
By defining
\begin{equation}
\mathbf{\Theta} = \mathbf{\Omega} \odot \mathbf{A},
\end{equation}
we obtain a $n\times n$ matrix~$\mathbf{\Theta}$ to characterize the mutual influence between neighboring vertices, where $\odot$ is the Hadamard product.
It's worth noting that $\mathbf{\Theta}$ is sparse due to the fact that $\mathbf{A}$ is sparse, and thus can be quickly computed.

Recall that each vertex has a one-hot vector to encode the 
probability in which the vertex belongs to each part. 
Let $v_i,v_j$ be a pair of neighboring vertices. 
For $v_i$, we extract the maximum component of its one-hot vector, denoted by $h_{i}$. Suppose that the maximum component occurs at the $s_i$-th slot. 
We then extract the corresponding component of the one-hot vector of $v_j$, denoted by $h_{j}^{s_i}$. Similarly, we can define $h_{j}$ and $h_{i}^{s_j}$.
We use $|h_{i}-h_{j}^{s_i}|+|h_{j}-h_{i}^{s_j}|$ to measure the symmetric label difference between $v_i$ and $v_j$.
In this way, we obtain a sparse matrix~$\mathbf{H}$ to characterize the pairwise label difference.
Therefore, the matrix~$\mathbf{\Theta} \odot \mathbf{H}$ is the improved difference matrix weighted by the above-mentioned influence. 
The overall label coherence can be obtained by summing the elements together:
\begin{equation}
\mathcal{L}_{\text{adj}} = \frac{1}{n} \cdot \mathbf{1}^\text{T} \times (\frac{\mathbf{\Theta} \odot \mathbf{H} \times \mathbf{1}}{\mathbf{A} \times \mathbf{1}}),
\end{equation}
where $\mathbf{1}$ is a column vector consisting of~$n$ $1$'s.

\section{Experiments}
\label{sec:experiments}

We present extensive experiments 
to validate the effectiveness 
on the mesh classification/segmentation tasks. 
The data preprocessing/augmentation step is conducted after SubdivNet~\cite{Hu2022SubdivNet}.
All the meshes are scaled into a unit cube. 
During the training phase, we apply an isotropic scaling operator on each model,
and the scaling factor is randomly sampled from a normal distribution with an expected value~$\xi=1$ and a standard deviation~$\sigma=0.1$. 
We also randomly select three Euler angles,
each of which being 0, or $\pi / 2$, or $\pi$, or $3\pi / 2$, to perform the orientation augmentation. 
The meshes in each dataset are simplified to have roughly equally many faces.

\subsection{Mesh Classification}
We demonstrate the superior classification ability of Laplacian2Mesh on the two datasets: SHREC11~\cite{Lian2011SHRECT} and Manifold40~\cite{Hu2022SubdivNet}. 
The classification accuracy statistics are reported to compare the performance among different methods.

\textbf{SHREC11.}
The SHREC11 dataset consists of 30 classes, with 20 examples per class. Following the setting of~\cite{Ezuz2017GWCNNAM}, all the methods are evaluated based on the 16-4 and 10-10 train-test split, respectively.
Our method outperforms the others on both train-test splits, and 
achieves the perfect classification accuracy ($100\%$), as shown in Table~\ref{tab:shrec11}. 
Tests on the SHREC11 dataset indicate that our Laplacian2Mesh is competitive on the mesh classification task.

\begin{table}[ht]
  \caption{The classification accuracy statistics on the SHREC11 dataset~\cite{Lian2011SHRECT}.}
  \label{tab:shrec11}
\begin{center}
  \begin{tabular}{%
	l%
	*{2}{c}%
	}
  \toprule
  \textbf{Method} & \textbf{Split 16} & \textbf{Split 10}  \\
  \midrule
  GWCNN~\cite{Ezuz2017GWCNNAM}              & 96.6$\%$ & 90.3$\%$ \\
  MeshCNN~\cite{Hanocka2019MeshCNNAN}       & 98.6$\%$ & 91.0$\%$ \\
  PD-MeshNet~\cite{Milano2020PrimalDualMC}  & 99.7$\%$ & 99.1$\%$ \\
  MeshWalker~\cite{Lahav2020MeshWalkerDM}   & 98.6$\%$ & 97.1$\%$ \\
  SubdivNet~\cite{Hu2022SubdivNet}          & 99.9$\%$ & 99.5$\%$ \\
  HodgeNet~\cite{Smirnov2021HodgeNetLS}     & 99.2$\%$ & 94.7$\%$ \\
  DiffusionNet (xyz)~\cite{Nicholas2022DiffusionNet} & - & 99.4$\%$ \\
  DiffusionNet (hks)~\cite{Nicholas2022DiffusionNet} & - & 99.5$\%$ \\
  Laplacian2Mesh (ours)     & \textbf{100}$\%$ & \textbf{100}$\%$ \\
  \bottomrule
  \end{tabular}%
  \end{center}
\end{table}

\begin{figure*}[tb]
 \begin{center}
 \includegraphics[width=\linewidth]{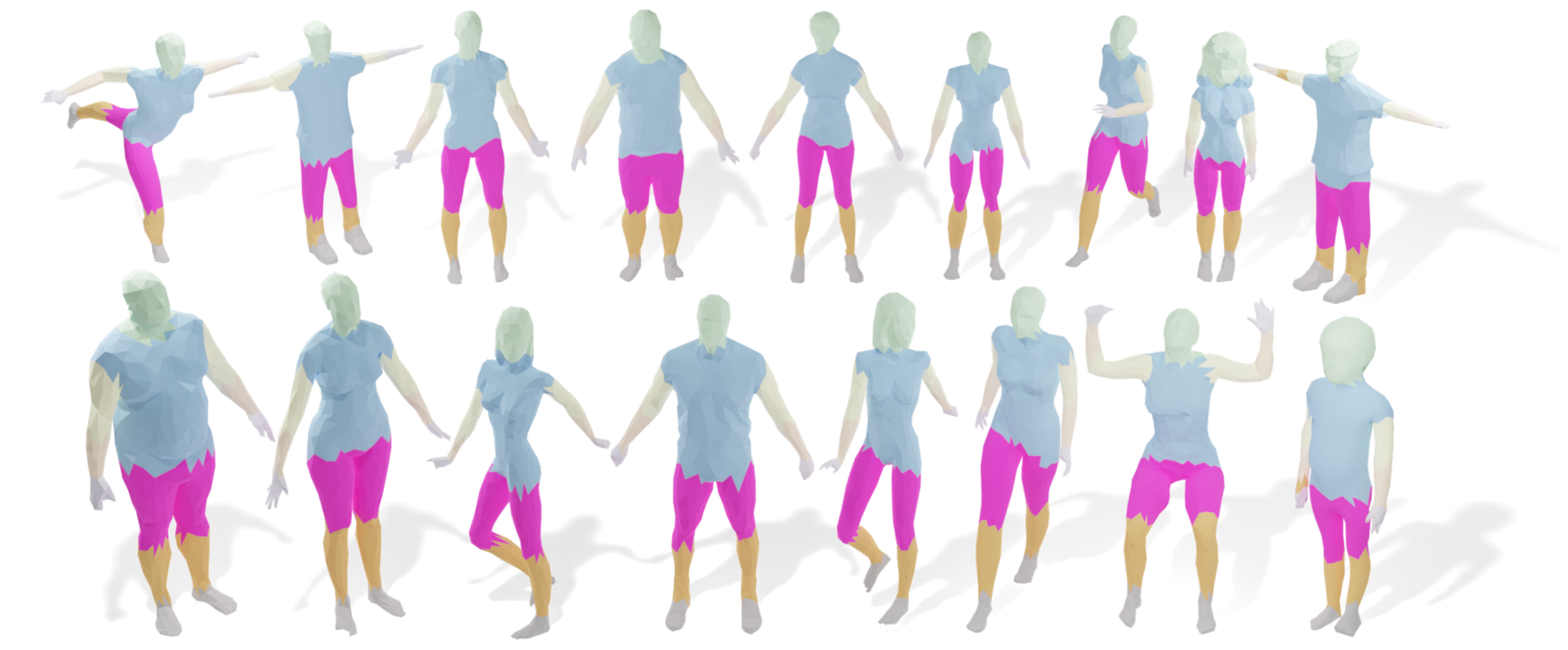}
 \end{center}
 \caption{We test the dataset of Human Body, and visualize the segmentation result for every model.
 }
 \label{fig:humanbody_seg}
\end{figure*}

\textbf{Manifold40.}
Manifold40 is a larger and more challenging dataset, containing 12311 CAD models from 40 categories. It is reconstructed from ModelNet40~\cite{Wu20153DSA} to obtain better triangulation quality meshes. Therefore, the reconstruction and simplification operations during the dataset construction stage 
inevitably introduce a slight difference in shape and a significant difference in tessellation. 
As shown in Table~\ref{tab:manifold40}, our method has a comparable performance to SubdivNet~\cite{Hu2022SubdivNet}, and gains better accuracy than other methods.
But we must point out that our approach does not include a tedious step of 
mesh subdivision~\cite{Loop1987SmoothSS} as included in SubdivNet~\cite{Hu2022SubdivNet}.

\begin{table}[ht]
  \caption{The classification accuracy statistics on Manifold40~\cite{Hu2022SubdivNet}.}
  \label{tab:manifold40}
\begin{center}
  \begin{tabular}{%
	l%
	c%
    c%
	}
  \toprule
  \textbf{Method} & \textbf{Input} & \textbf{Accuracy} \\
  \midrule
  PointNet++~\cite{Qi2017PointNetDH}        & point cloud & 87.9$\%$ \\
  PCT~\cite{Guo2021PCTPC}                   & point cloud & \textbf{92.4}$\%$ \\
  \midrule
  MeshNet~\cite{feng2019meshnet}            & mesh & 88.4$\%$ \\
  MeshWalker~\cite{Lahav2020MeshWalkerDM}   & mesh & 90.5$\%$ \\
  SubdivNet~\cite{Hu2022SubdivNet}          & mesh & \textbf{91.2}$\%$ \\
  Laplacian2Mesh (ours)                     & mesh & 90.9$\%$ \\
  \bottomrule
  \end{tabular}%
  \end{center}
\end{table}

\subsection{Mesh Semantic Segmentation}

We conduct the mesh semantic segmentation experiments on the human body dataset~\cite{Maron2017ConvolutionalNN} and COSEG dataset~\cite{Wang2012ActiveCO}. The mesh segmentation task aims to predict the segmentation label per face. Since the labels are defined on the mesh vertices in our approach, we apply the majority voting to obtain the labels on the faces, rather than the ``soft'' label conversion or some smooth processing technique~\cite{Lahav2020MeshWalkerDM}. 

\textbf{Human Body Segmentation.}
The human body dataset, labeled by~\cite{Maron2017ConvolutionalNN}, consists of 370 training shapes from SCAPE~\cite{Anguelov2005SCAPESC}, FAUST~\cite{Bogo2014FAUSTDA}, MIT~\cite{Vlasic2008ArticulatedMA}, and Adobe Fuse~\cite{Adobe16}. The 18 models in the test set are from SHREC07~\cite{giorgi2007shape} (humans) dataset. All the meshes are manually segmented into 8 parts~\cite{Kalogerakis2010Learning3M}. We use the same version of human body dataset as in MeshCNN~\cite{Hanocka2019MeshCNNAN}, which is downsampled to 1500 faces per mesh.

As shown in Figure~\ref{fig:humanbody_seg}, our method is able to learn the consistent and accurate part segmentation of human bodies. We report the segmentation performance of various methods in Table~\ref{tab:humanbody}. Our method mostly outperforms the related works such as MeshCNN~\cite{Hanocka2019MeshCNNAN}, PD-MeshNet~\cite{Milano2020PrimalDualMC}, and HodgeNet~\cite{Smirnov2021HodgeNetLS}. Despite being slightly inferior to the DiffusionNet~\cite{Nicholas2022DiffusionNet}, 
our method is superior in terms of its noise-resistance property, as will be demonstrated in Section~\ref{sec:Resistance2noise}. 

\begin{table}[ht]
  \caption{The mesh segmentation accuracy statistics on the Human-Body dataset~\cite{Maron2017ConvolutionalNN}. 
  Note that DiffusionNet~\cite{Nicholas2022DiffusionNet} supports various inputs.
  The options ``xyz'' and ``hks'' denote the raw coordinates and the heat kernel signatures, respectively.
  }
  \label{tab:humanbody}
\begin{center}
  \begin{tabular}{lcc}
  \toprule
  \textbf{Method} & \textbf{Input} & \textbf{Accuracy} \\
  \midrule
  PointNet~\cite{Qi2017PointNetDL}          & point cloud & 74.7$\%$ \\
  PointNet++~\cite{Qi2017PointNetDH}        & point cloud & 82.3$\%$ \\
  \midrule
  MeshCNN~\cite{Hanocka2019MeshCNNAN}       & mesh & 85.4$\%$ \\
  PD-MeshNet~\cite{Milano2020PrimalDualMC}  & mesh & 85.6$\%$ \\
  HodgeNet~\cite{Smirnov2021HodgeNetLS}     & mesh & 85.0$\%$ \\
  DiffusionNet (xyz)~\cite{Nicholas2022DiffusionNet} & mesh & 88.8$\%$ \\
  DiffusionNet (hks)~\cite{Nicholas2022DiffusionNet} & mesh & \textbf{90.5}$\%$ \\
  Laplacian2Mesh (ours)                     & mesh & 88.6$\%$ \\
  \bottomrule
  \end{tabular}%
  \end{center}

\end{table}

\begin{figure}[tb]
 \begin{center}
 \includegraphics[width=\columnwidth]{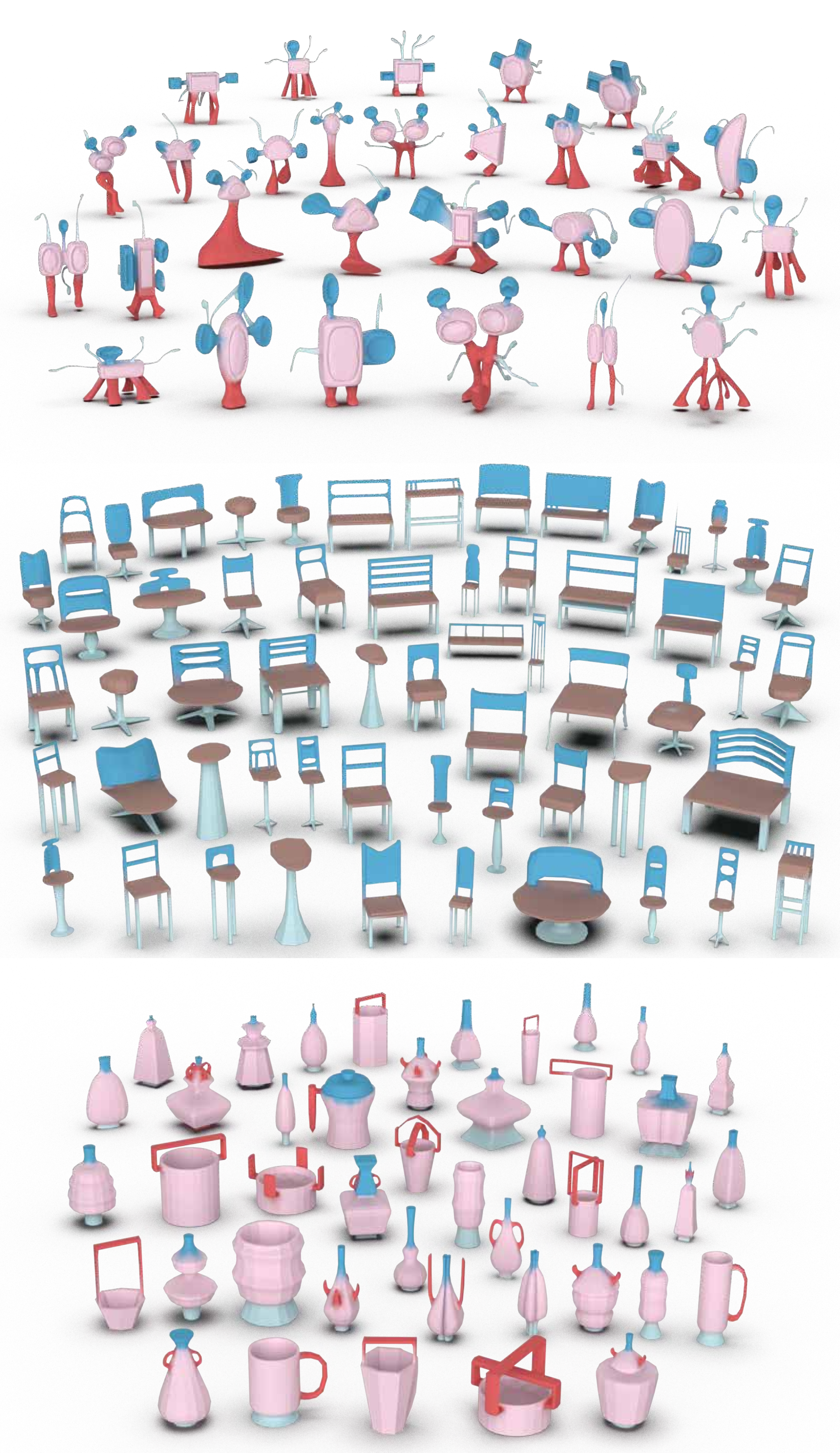}
 \end{center}
 \caption{A gallery of segmentation results of the COSEG dataset. From top to bottom: Tele-aliens, Chairs, and Vases classes.}
 \label{fig:coseg}
\end{figure}

\textbf{COSEG.}
We also evaluate the performance of Laplacian2Mesh on the three largest sets from the COSEG shape dataset: Vases, Chairs, and Tele-aliens containing 300, 400, and 200 models, respectively. The meshes are labeled into 3 parts (Chairs set) or 4 parts (Vases set $\&$ Tele-aliens set). We generate a random 85$\%$-15$\%$ train-test split for each set, as in MeshCNN~\cite{Hanocka2019MeshCNNAN}. 

The quantitative results are provided in Table~\ref{tab:coseg}, and the qualitative results for all models in the test set are visualized in Figure~\ref{fig:coseg}. Our method outperforms  other methods on the Vase and Chair classes, but performs worse on the Tele-alien class.
We explain this as follows.
The triangulation quality of the Tele-alien meshes is bad, containing long and thin triangles. To our knowledge, the Laplacian matrix is likely to be ill-conditioned when bad triangles exist.
Figure~\ref{fig:coseg} visualizes the semantic segmentation results computed by our method.

\begin{table}[tb]
  \caption{The mesh segmentation accuracy statistics on the COSEG dataset~\cite{Wang2012ActiveCO}.}
  \label{tab:coseg}
\begin{center}
  \setlength{\tabcolsep}{1mm}{
  \begin{tabular}{%
	l%
	*{3}{c}%
	}
  \toprule
  \textbf{Method} & \textbf{Vases} & \textbf{Chairs} & \textbf{Tele-aliens} \\
  \midrule
  DCN~\cite{Xu2017DCN}                  & 90.9$\%$ & 95.7$\%$ & - \\
  MeshCNN~\cite{Hanocka2019MeshCNNAN}   & 92.4$\%$ & 93.0$\%$ & \textbf{96.3}$\%$ \\
  HodgeNet~\cite{Smirnov2021HodgeNetLS} & 90.3$\%$ & 95.7$\%$ & 96.0$\%$ \\
  Laplacian2Mesh (ours) & \textbf{94.6}$\%$ & \textbf{96.6}$\%$ & 95.0$\%$ \\
  \bottomrule
  \end{tabular}}%
  \end{center}
\end{table}

\subsection{Ablation Studies}
\label{subsec:ablation_studies}

The ingredients of our method lie in many aspects, including the selection of the input features, the multi-resolution spectral signals, the network architecture, the adjacency loss, and the selection of the kernel size for the training. We evaluate our method by conducting the ablation studies on the mesh segmentation task.

\textbf{The selection of the input features.}
As described in Section~\ref{sec:input_features}, we pre-compute a set of intrinsic and extrinsic geometric features and transform them into the spectral signals as the input of our network. Table~\ref{tab:ablation_features} reports the segmentation performance on the Vase class of COSEG dataset, Manifold40 dataset, and the human body dataset after removing each of the features. It shows that the Laplacian2Mesh with the full set of features achieves the best performance on all the three datasets. In Table~\ref{tab:ablation_features}, we highlight the features that cause the greatest performance degradation on each dataset. It indicates that various classes have different dependency on the input features.
Our Laplacian2Mesh can automatically learn the weighting scheme to fuse the diverse geometric features.

\begin{table}[t]
  \caption{The results of ablation experiments on the input features. We highlight the best (bold) and the worst (underlining) performance scores for each class.
  It can be seen that the importance of geometric features
  varies on different datasets.
  }
  \label{tab:ablation_features}
  \begin{center}
  \setlength{\tabcolsep}{1mm}{
  \begin{tabular}{%
	l%
	*{3}{c}%
	}
  \toprule
  \textbf{Method} & \textbf{Vases} & \textbf{Manifold40} & \textbf{Human Body} \\
  \midrule
  w/o vertex coordinates     & \underline{85.9$\%$} & 88.7$\%$ & 87.4$\%$ \\
  w/o vertex normal          & 93.2$\%$ & 90.2$\%$ & 87.3$\%$ \\
  w/o Gaussian Curvature     & 92.2$\%$ & 90.3$\%$ & 87.7$\%$ \\
  w/o EigenVectors & 86.3$\%$ & \underline{85.3$\%$} & 88.2$\%$ \\
  w/o HKS                    & 92.0$\%$ & 89.8$\%$ & \underline{77.5$\%$} \\
  w/o Dihedral Angles        & 90.5$\%$ & 90.1$\%$ & 87.1$\%$ \\
  Laplacian2Mesh (ours)      & \textbf{94.6}$\%$ & \textbf{90.9}$\%$ & \textbf{88.6}$\%$ \\
  \bottomrule
  \end{tabular}}%
  \end{center}

\end{table}

\begin{table}[t]
  \caption{The segmentation performance with the input spectral signals of different resolutions. The first three rows are the single-resolution experiments, indicating that it is proper to take $k$ as one half of the total number of vertices. The other rows show the results of using the multi-resolution inputs.}
  \label{tab:ablation_input}
\begin{center}
  \begin{tabular}{%
	l%
	c%
	c%
	}
 \toprule
 \textbf{input sizes} & \textbf{Human Body} & \textbf{Tele-alien} \\
  \midrule
   749-0-0 & 84.8$\%$ & 90.7$\%$ \\
   512-0-0 & 86.9$\%$ & 93.1$\%$ \\
   128-0-0 & 86.7$\%$ & 92.9$\%$ \\
  512-128-0 & 87.1$\%$ & 93.3$\%$ \\
   512-128-32 & \textbf{88.6$\%$} & 94.5$\%$ \\
   512-256-64 & 86.3$\%$ & \textbf{95.0$\%$} \\
  \bottomrule
  \end{tabular}%
  \end{center}
\end{table}

\textbf{The multi-resolution signals.}
As described in Section~\ref{sec:spectral_transform}, the selection of the hyperparameter $k$ is of vital importance to form the spectral basis, which 
helps filter out the high-frequency noise. We compute the multi-resolution spectral signals as the input of our network by selecting $\mathbf{k}=\{k_i \mid i=0,1,2\}$ to form a set of basis and perform the spectral transform.

Table~\ref{tab:ablation_input} reports the segmentation performance statistics with different hyperparameter settings, including the single-resolution experiments (e.g. $128-0-0$, $512-0-0$, $749-0-0$, with $749$ being the smallest vertex number among all the meshes in the dataset) and the multi-resolution experiments (e.g. $512-128-0$, $512-128-32$, $512-256-64$). 
By comparing the single-resolution experiments, it shows the
importance of choosing a proper $k$, as either the largest or
a smaller $k$ will decrease the performance. On the other
hand, the comparison between the single-resolution and
multi-resolution experiments validates the effectiveness of
our setting.
 
\begin{figure}[tb]
 \begin{center}
 \includegraphics[width=\columnwidth]{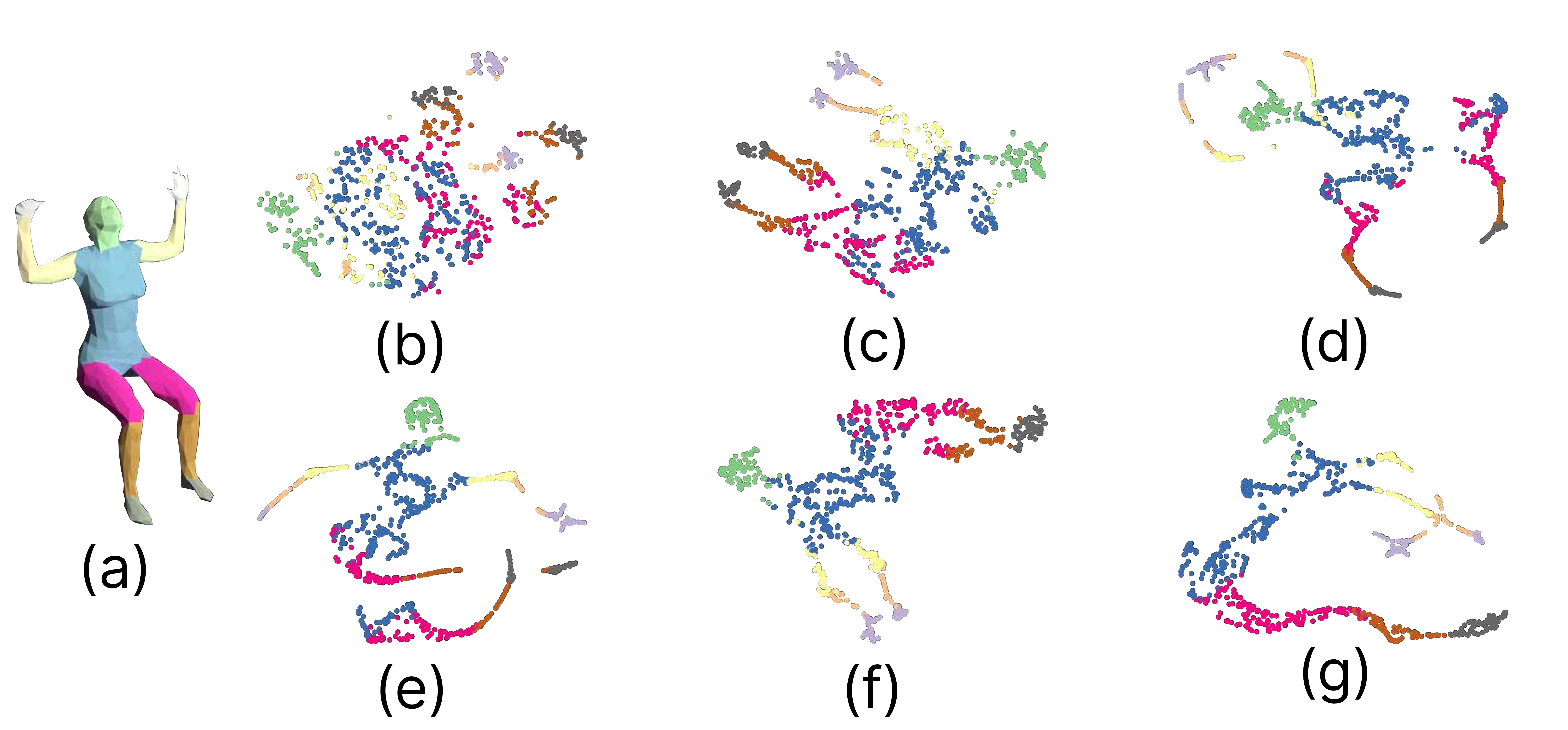}
 \end{center}
 \caption{
 t-SNE visualization of the network processing. We project the intermediate spectral features back to the spatial vertex-wise features for a better understanding. The vertices are colored by their ground-truth segmentation labels. (a): the ground-truth segmentation; (b-d): the encoded feature of each resolution before any Laplacian pooling/unpooling; (e): the encoded feature after fusing all the resolutions; (f): the decoded feature after the U-Net processing, (g): the output segmentation predictions.
 }
 \label{fig:tsne_internal_layers}
\end{figure}

From a different perspective, it is intuitive and interesting to visualize the learned features to understand the role of the multi-resolution input signals. Note that the whole network processing is in the Laplacian spectral domain, and therefore we need to transform the spectral features $\widetilde{\mathbf{F}}$ back to the spatial domain via the inverse mapping:
\begin{equation}
\label{equ:trans2Euclidean}
\widehat{\mathbf{F}}=\mathbf{\Phi}_{k} \widetilde{\mathbf{F}}.
\end{equation}
We show the t-SNE visualization of the intermediate features in Figure~\ref{fig:tsne_internal_layers}, where the color indicates the ground-truth label of each vertex. The first row is the encoded feature of each resolution before any pooling/unpooling operation. The signals with mostly low-frequency signals, i.e. $k=32$, obtain more disjoint clusters, while the signals with more high-frequency signals, i.e. $k=512$ have the encoded features gathered together probably because of the distracting local details. However, after we fuse the features from the multi-resolution signals during encoding, the clusters are more separated at the boundary with the help of the local information (see (d) and (e)). Finally, the features are moved and clustered based on their semantic labels, as the vertices on the human legs are moved together after the decoding stage (see (e) and (f)). This visualization shows the necessity of fusing the multi-resolution signals.

\begin{figure}[tb]
 \begin{center}
 \includegraphics[width=\columnwidth]{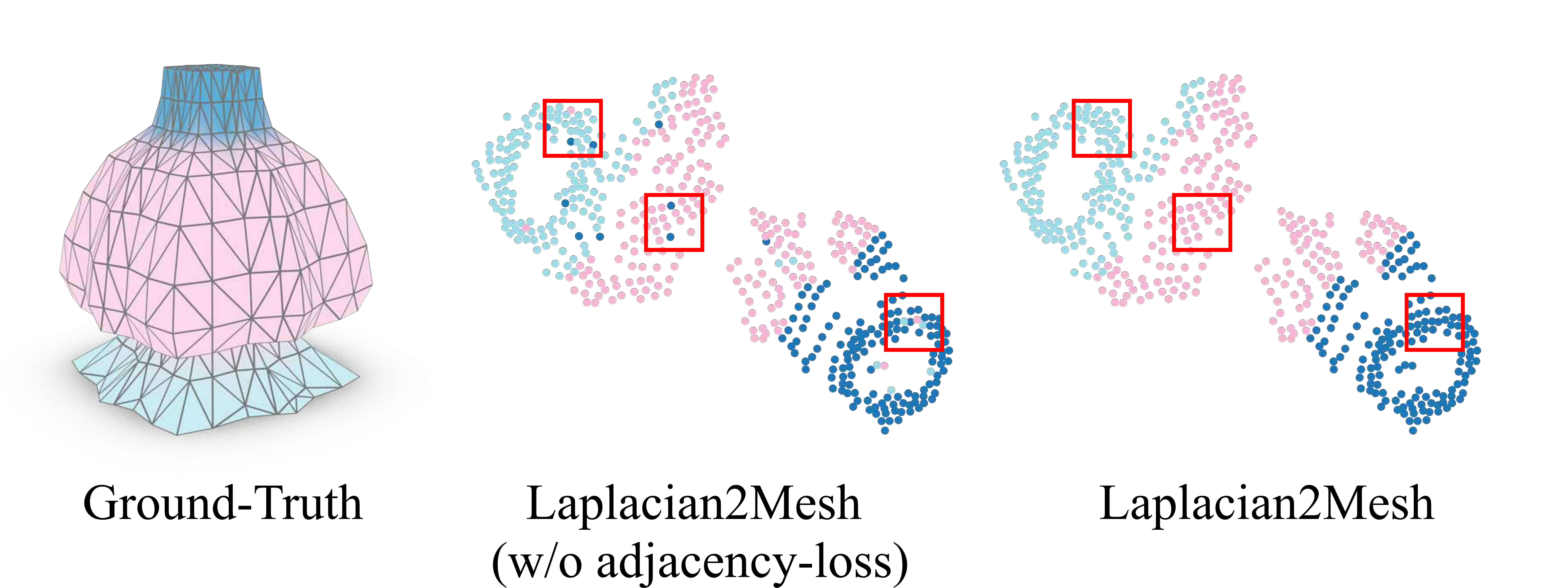}
 \end{center}
 \caption{t-SNE visualization of the segmentation results with and without the adjacency loss. The vertices are colored by their ground-truth label. The adjacency loss obviously improves the smoothness of the segmentation results.}
 \label{fig:adj_loss}
\end{figure}

\begin{table}[tb]
  \caption{The ablation studies of the network architecture and the adjacency loss. }
  \label{tab:ablation_network}
\begin{center}
  \begin{tabular}{%
	l%
	c%
	c%
	}
  \toprule
 \textbf{Experiments} & \textbf{Vases} & \textbf{Chairs} \\
  \midrule
   Baseline & 89.2$\%$ & 93.8$\%$ \\
   + SE-block & 90.6$\%$ & 94.9$\%$ \\
   + Multi-resolution & 91.3$\%$ & 95.5$\%$ \\
   + Laplacian pooling/unpooling & 93.3$\%$ & 96.3$\%$ \\
  + Adjacency loss (full) & \textbf{94.6}$\%$ & \textbf{96.6}$\%$ \\
  \bottomrule
  \end{tabular}%
  \end{center}

\end{table}

\textbf{Network Architecture and Loss.} 
We show how it evolves with the key designs in our network and the loss function. The quantitative evaluations are in Table~\ref{tab:ablation_network}. We start from a baseline network implemented as a simple convolutional network with skip connections, which is similar to the one-resolution component of our network while the SE-blocks are replaced by the vanilla convolutional blocks. We progressively equip the network with the SE-blocks with small-sized convolution kernels, the multi-resolution network architecture but with spatial pooling/unpooling, our Laplacian pooling/unpooling operations, and the adjacency loss. The quantitative evaluation clearly demonstrates that our key designs are necessary to learn the mesh semantics in the spectral domain.

The adjacency loss is used to guarantee the label coherence property of the semantic segmentation on the mesh surface. As indicated by the t-SNE visualization shown in Figure ~\ref{fig:adj_loss}, without the adjacency loss, some outlier vertices may be assigned with a wrong label, 
leading to conspicuous visual segmentation artifacts. 

\textbf{Kernel Size.}
As said before, our network should be equipped with small kernel convolution. As illustrated in Figure~\ref{fig:se_resnet}, we directly use the common convolution kernel settings for SE-block~\cite{He2018BagOT}, which is $ks=3$. We also tried using other kernel sizes for SE-block, but $ks=3$ got the best results. For the layers without the SE-block (the MLPs in our network), we experimented with the impact of different sizes of the kernel on the segmentation results.

We replace the MLPs in our network with convolutions of different kernel sizes (i.e. 1, 3, 5, 7). From Table~\ref{tab:kernel_size}, we can also find that a large convolution kernel does not bring us similar benefits as it is used in 2D CNNs, on the contrary, the large-sized kernel will degrade the performance of our network. So we select $ks=1$ (i.e. MLP) for the layers without the SE-block.

\renewcommand\arraystretch{1.2}
\begin{table}[ht]
  \caption{The segmentation performance with the different kernel sizes.}
  \label{tab:kernel_size}
  \begin{center}
  {
  \begin{tabular}{%
	l%
	*{3}{c}%
	}
  \toprule
    & \textbf{Vases} & \textbf{Chairs} & \textbf{Human Body} \\
  \midrule
  kernel\_size = 7  & 92.3$\%$ & 94.8$\%$ & 86.5$\%$ \\
  kernel\_size = 5  & 92.8$\%$ & 95.6$\%$ & 87.1$\%$ \\
  kernel\_size = 3  & 93.5$\%$ & 96.1$\%$ & 87.5$\%$ \\
  kernel\_size = 1  & \textbf{94.6}$\%$ & \textbf{96.6}$\%$ & \textbf{88.6}$\%$ \\
  \bottomrule
  \end{tabular}}%
  \end{center}
\end{table}

\section{Strengths and Limitations}
\label{sec:limitations}

\begin{figure}[tb]
 \begin{center}
 \includegraphics[width=\linewidth]{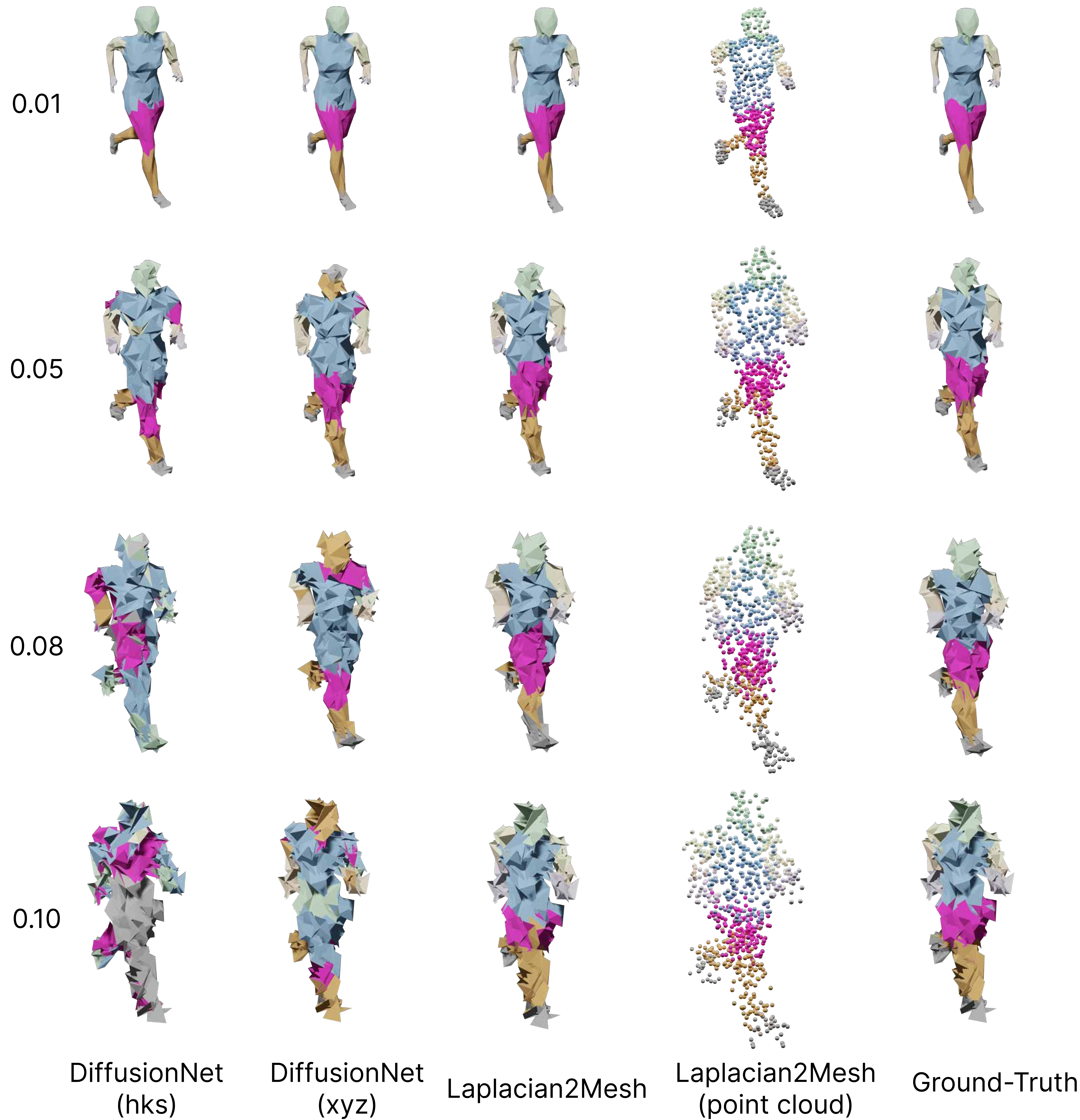}
 \end{center}
 \caption{
 The qualitative results of testing the noise-resistance. 
 From top to bottom: As noise levels rise, segmentation outputs of DiffusionNet diverge significantly from the ground-truth, while ours is more in line with it. 
 }
 \label{fig:anti-noise}
\end{figure}

\subsection{Resistance to Noise}
\label{sec:Resistance2noise}

To test the robustness against noise, we construct the noisy dataset by adding Gaussian noise of several levels (i.e. 0.005, 0.010, 0.050, 0.080, 0.1 w.r.t. the diagonal length of each model) to the original human body dataset~\cite{Maron2017ConvolutionalNN}. We train our network with the original noise-free dataset and test on the noisy meshes. Figure~\ref{fig:anti-noise} shows the comparison between DiffusionNet and ours with various settings.

Table~\ref{tab:anti-noise} reports the quantitative comparisons.
The DiffusionNet performance has significantly decreased, as can be seen, whereas our performance has essentially been steady. 
Interestingly, once we replace the Laplacian pooling/unpooling of our network by the max pooling (in other words, we apply the spatial U-Net directly to our spectral signals), the performance has an obvious drop.
This indicates the necessity of our design to deal with the spectral signals.

\begin{table}[tb]
  \caption{The mesh segmentation accuracy statistics when training on the original human body dataset~\cite{Maron2017ConvolutionalNN} and testing on the noisy meshes. $\clubsuit$ means that we use the max-pooling instead of Laplacian pooling in our network. The DiffusionNet has a drastic decrease while ours does not, which shows that ours is more robust to noise.}
  \label{tab:anti-noise}
\begin{center}
  \begin{tabular}{lcc}
  \toprule
  \textbf{Method} & \textbf{w/ Noise} & \textbf{w/o Noise} \\
  \midrule
  DiffusionNet (xyz)~\cite{Nicholas2022DiffusionNet} & 64.4$\%$ & 88.8$\%$ \\
  DiffusionNet (hks)~\cite{Nicholas2022DiffusionNet} & 62.0$\%$ & \textbf{90.5}$\%$ \\
  Laplacian2Mesh (ours) $^{\clubsuit}$    & 60.2$\%$ & 85.9$\%$ \\
  Laplacian2Mesh (ours)   & \textbf{86.7}$\%$  & 88.6$\%$\\
  \bottomrule
  \end{tabular}%
  \end{center}
\end{table}

\subsection{Non-watertight $\&$ Non-manifold Mesh Segmentation}

\begin{figure}[tb]
 \begin{center}
 \includegraphics[width=\columnwidth]{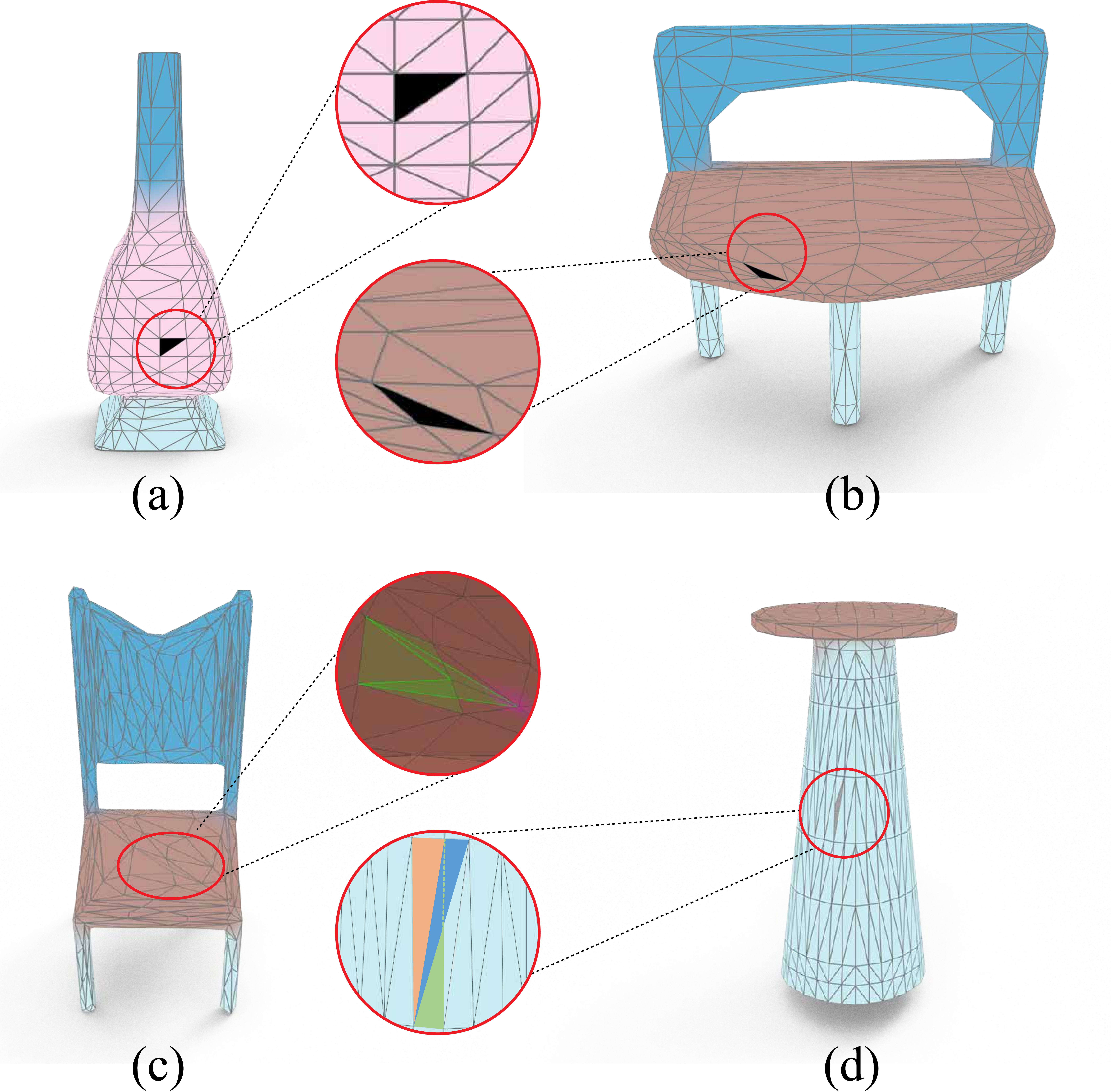}
 \end{center}
 \caption{Our Laplacian2Mesh is able to perform segmentation on the non-watertight and non-manifold data. (a) and (b) are non-watertight vases and chairs, respectively; (c) has a non-manifold vertex, and (d) has a non-manifold edge.}
 \label{fig:unclosed_nonmanifold}
\end{figure}

Many mesh segmentation methods~\cite{Hanocka2019MeshCNNAN, Hu2022SubdivNet} are designed on the premise that the input mesh must be watertight or manifold or both, which limits their usage on a wider range of various mesh datasets. For example, SubdivNet~\cite{Hu2022SubdivNet} assumes that every face of the mesh should have 3 neighbor faces. MeshCNN~\cite{Hanocka2019MeshCNNAN} requires each edge to be shared by two triangular faces. Therefore, these methods will fail even with a small number of small gaps or non-manifold elements (non-manifold vertices or non-manifold edges). 

We show our segmentation results on the non-watertight and non-manifold meshes in Figure~\ref{fig:unclosed_nonmanifold}. Our Laplacian2Mesh is flexible to deal with various structures~\cite{Pinkall1993ComputingDM, Sharp2020ALF}. If the edge is a non-manifold edge, our solution is to use robust-laplacian~\cite{Sharp2020ALF} and discard all dihedral angles corresponding to this edge. In addition, if the face $\mathbf{f}_i$ is a boundary face, we will fill the missing dihedral angle value with the value 1, which means that this location has gentle terrain.

\begin{figure}[ht]
 \centering
 \includegraphics[width=\columnwidth]{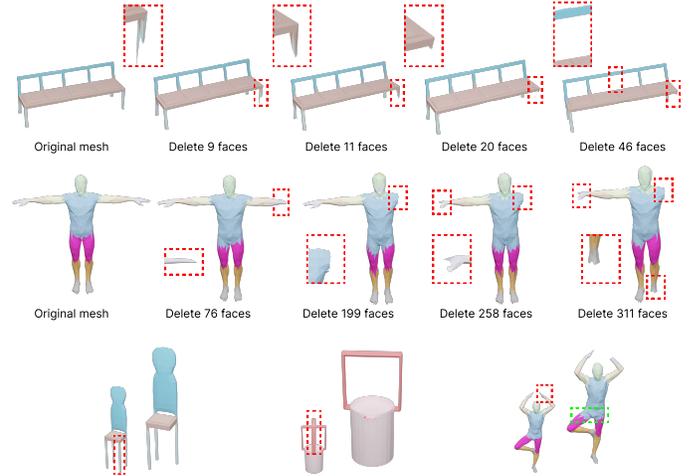}
 \caption{The results on incomplete models. The red dashed boxes on the mesh indicate the deleted parts. The green dashed boxes highlight the inaccurate segmentation boundaries. }
 \label{fig:incomplete_mesh}
\end{figure}
\vspace{1em}

\subsection{Incomplete Models}

We select some models from the COSEG dataset and the Human-Body dataset, and manually delete some triangle faces. Note that it results in open surfaces, which cannot be processed by other mesh understanding networks. We test these incomplete models with the pre-trained segmentation network.

The top two rows of Figure~\ref{fig:incomplete_mesh} show the segmentation results when more and more triangle faces are removed, while the last row shows the segmentation of different shapes. It can be observed that the overall segmentation results remain reasonable and sound, indicating that our model can be generalized to incomplete models. However, as shown in the human case, the broken models may have poorly fine-grained segmentation at the boundaries of the parts near the removed region. It's an interesting future direction to study that how the spectral-based deep understanding methods are affected by different geometric and topological changes.

\subsection{More Merits}
Besides the ability to deal with all kinds of mesh defects and noise,
the biggest advantage of our approach is to decouple various shape understanding tasks from the tedious and complicated triangulation. 
It can also support various kinds of inputs, such as meshes and point clouds,
as long as the Laplacian matrix can be equipped. 
To summarize, learning the latent features in the spectral domain rather than the spatial domain inherits the nice properties of the spectral analysis.

\subsection{Limitations}

\begin{figure}[tb]
 \begin{center}
 \includegraphics[width=\columnwidth]{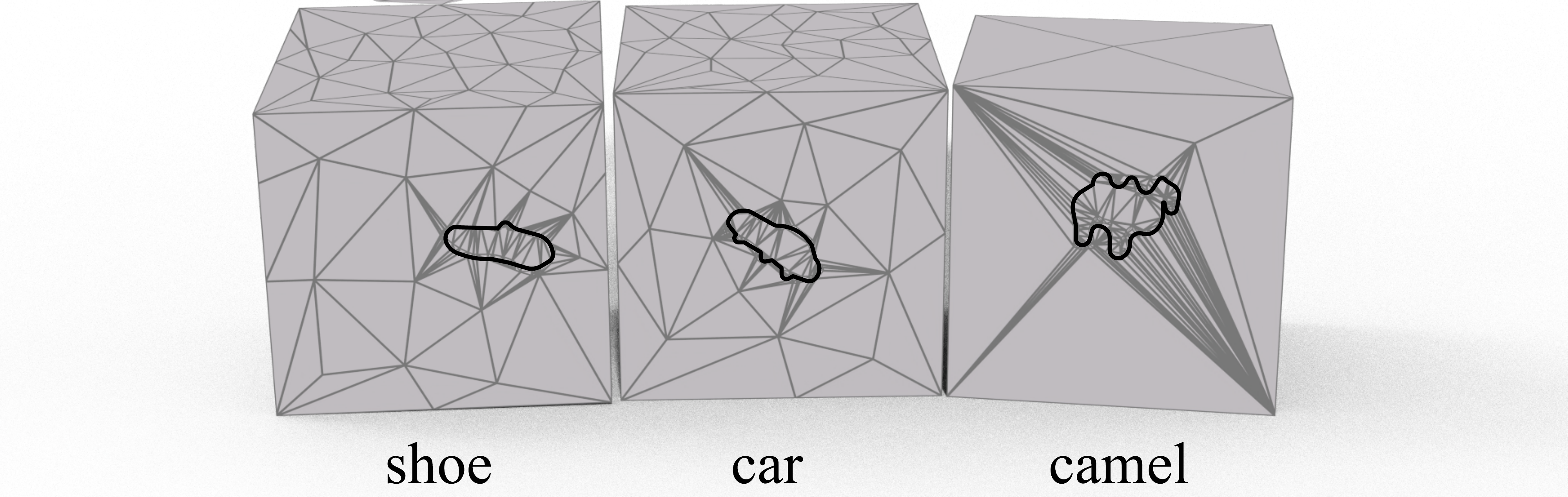}
 \end{center}
 \caption{
 The Cube Engraving dataset~\cite{Latecki2000ShapeSM}. Our method fails on this dataset since our approach deems the engraved objects as high-frequency information.
 }
 \label{fig:limitation}
\end{figure}

The side-effect of the high-frequency filtering in our spectral transform is the missing of local features in some unusual cases. Taking the Cube Engraving dataset~\cite{Latecki2000ShapeSM} as an example, each model in this dataset is a cube with a facet being ``engraved'', as demonstrated in Figure~\ref{fig:limitation}.
The engraved objects are important semantic hints. 
This dataset contains 4600 objects with the 3910-690 train/test split.
But our approach regards the engraved objects as high-frequency information.
Not surprisingly, as reported in Table~\ref{tab:cubes}, our Laplacian2Mesh ignores the local shape classification and performs worst among all the mesh classification methods.

\renewcommand\arraystretch{1}
\begin{table}[ht]
  \caption{The classification accuracy statistics on the Cube Engraving dataset~\cite{Latecki2000ShapeSM}. }
  \label{tab:cubes}
\begin{center}
  \begin{tabular}{%
	l%
	c%
	}
  \toprule
  \textbf{Method} & \textbf{Accuracy} \\
  \midrule
  PointNet++~\cite{Qi2017PointNetDH}        & 64.3$\%$ \\
  MeshCNN~\cite{Hanocka2019MeshCNNAN}       & 92.2$\%$ \\
  PD-MeshNet~\cite{Milano2020PrimalDualMC}  & 94.4$\%$ \\
  MeshWalker~\cite{Lahav2020MeshWalkerDM}   & 98.6$\%$ \\
  SubdivNet~\cite{Hu2022SubdivNet}          & \textbf{98.9}$\%$ \\
  Laplacian2Mesh (ours)                     & 91.5$\%$  \\
  \bottomrule
  \end{tabular}%
  \end{center}

\end{table}

\section{Conclusion and Future Work}
\label{sec:conclusion}

We present Laplacian2Mesh, a general network architecture for mesh understanding in the spectral domain. Our network takes the multi-resolution spectral signals as the input, and follows the structure of the U-Net architecture. We design the small-sized SE-blocks and propose the Laplacian pooling/unpooling operations to fuse the features from different levels of resolutions. The ablation studies demonstrate the necessity of our key designs to enable the learning of spectral mesh signals. Compared to state-of-the-art methods, our approach not only achieves competitive or even better performances on the mesh classification and segmentation tasks, but also can handle non-watertight and non-manifold meshes.
It also owns the nice feature of being noise-resistant. 

There is a vast of directions for further exploration. First, to perform the Laplacian spectral transform, we need to compute the eigendecomposition of the Laplacian matrix for each shape. Consequently, the eigenvectors, or to put it another way, the spectral basis, are not aligned among the shapes in the dataset. In our Laplacian2Mesh network, we seek for the small-sized SE-blocks to solve this problem. However, aligning the spectral basis will enable us to make use of the larger receptive fields, which will probably improve the performance in the mesh understanding tasks. Also, although we have concluded that the multi-resolution spectral signals are stronger than the single-resolution input, it's still a remaining question on how to automatically select the hyper-parameters, i.e. the number of eigenvectors, to adapt to different shape datasets.

\appendices
\section{Eigendecomposition Matrix}
\label{app:cotangent_laplacian}

For triangle meshes, a common choice of the discrete Laplacian is cotangent Laplacian, which forms the basis for the theory of discrete holomorphic functions and discrete Riemann surfaces. 
\setlength{\columnsep}{1em}
\setlength{\intextsep}{0em}
\begin{wrapfigure}{r}{110pt}
   \vspace{-0.1\baselineskip}
   \includegraphics[width=110pt]{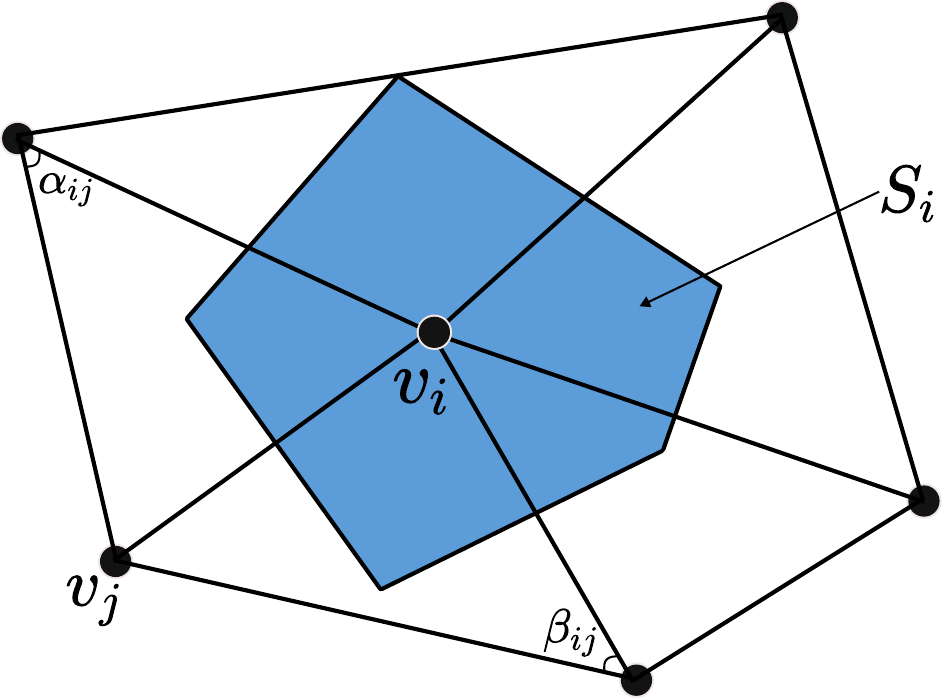}
   \vspace*{-1.5\baselineskip}
\end{wrapfigure}
It arises naturally via linear finite elements, discrete exterior calculus, electrical networks, and minimal surfaces. This operator is very sparse, easy to build, and generally works well for unstructured meshes with irregular vertex distributions, which also can be used on non-manifold meshes by accumulating per-triangle matrices~\cite{Pinkall1993ComputingDM, Sharp2020ALF}.

For a manifold triangle mesh, the discrete Laplacian $\Delta$ of a scalar function $f$ at a vertex $v_i$ can be approximated as:
\begin{equation}
\label{equ:cotangent_Laplacian_vertex}
\Delta f(v_i) := \frac{1}{2S_i} \sum_{v_j \in r_1(v_i)}{(cot \ \alpha_{ij} + cot \ \beta_{ij})(f(v_j)-f(v_i))}
\end{equation}

where $r_1(v_i)$ is over all the 1-ring adjacent vertices of $v_i$, $\alpha_{ij}$ and $\beta_{ij}$ are the two angles opposite to edge $v_iv_j$, and $S_i$ is the vertex area of $v_i$.

Then, the Laplacian is encoded in a sparse matrix $\mathbf{L} \in \mathbb{R}^{n \times n}$ for the mesh $\mathcal{M}$, such that
\begin{equation}
\label{equ:cotangent_Laplacian_mesh}
\begin{pmatrix} \vdots \\ \Delta f(v_i) \\ \vdots \end{pmatrix}
=
\mathbf{L} \cdot
\begin{pmatrix} \vdots \\ f(v_i) \\ \vdots \end{pmatrix}
\end{equation}

We define $\mathbf{M}$ as the diagonal matrix whose $i$-th entry along the diagonal is twice the vertex area $S_i$, that is
\begin{equation}
\label{equ:mass_matrix}
\mathbf{M}^{-1} = diag(\cdots, \frac{1}{2S_i}, \cdots).
\end{equation}
Then, the discretization of the Laplacian can be represented as
\begin{equation}
\label{equ:discretization_laplacian}
\mathbf{L} = \mathbf{M}^{-1}\mathbf{C}.
\end{equation}

As we know from~\cite{Lehoucq1998ARPACKUG}, the eigendecomposition of $\mathbf{L}$ is the generalized eigenvalue problem for $\mathbf{\lambda}$ eigenvalues with corresponding eigenvectors $\Phi$, which satisfies the following formula:
\begin{equation}
\label{equ:generalized_eigenvalue}
\mathbf{L} \mathbf{\Phi} = \mathbf{M}^{-1}\boldsymbol{\lambda}\mathbf{\Phi},
\end{equation}
where $\boldsymbol{\lambda}=diag(\lambda_0, \lambda_1, \cdots, \lambda_{n-1})$, and $\mathbf{\Phi}^T \mathbf{M}^{-1} \mathbf{\Phi}=\mathbf{I}$.

To facilitate computation, the generalized eigenvalue problem is equivalently transformed into the standard eigenvalue problem. So Eq.~\ref{equ:generalized_eigenvalue} is rewritten as
\begin{equation}
\label{equ:standard_eigenvalue}
(\mathbf{M} \mathbf{L}) \mathbf{\Phi} = \boldsymbol{\lambda}\mathbf{\Phi}.
\end{equation}
Deriving from Eq.~\ref{equ:discretization_laplacian}, we know that Eq.~\ref{equ:standard_eigenvalue} can be written as
\begin{equation}
\label{equ:cot_weight}
\mathbf{C} \mathbf{\Phi} = \boldsymbol{\lambda}\mathbf{\Phi},
\end{equation}
where $\mathbf{\Phi}^T \mathbf{\Phi}=\mathbf{I}$ in the formula.

To simplify computations in our work, the eigenvector matrix $\mathbf{\Phi}$ is derived from performing an eigendecomposition on the cotangent weight matrix (Eq.~\ref{equ:cot_weight}) instead of the discrete Laplace-Beltrummy matrix (Eq.~\ref{equ:generalized_eigenvalue}).

In~\cite{Lvy2006LaplaceBeltramiET}, since the authors obtained the eigenvectors by eigendecomposition of Laplacian, the spectral reconstruction is written as
\begin{equation}
\label{equ:spectral_reconstruction_laplacian}
\widehat{\mathbf{V}}=\mathbf{\Phi}_{k} (\mathbf{\Phi}_{k}^T \mathbf{M} \mathbf{V}).
\end{equation}
If we directly eigendecompose the cotangent weight matrix, then the spectral reconstruction is
\begin{equation}
\label{equ:spectral_reconstruction_cotangent}
\widehat{\mathbf{V}}=\mathbf{\Phi}_{k} (\mathbf{\Phi}_{k}^T \mathbf{V}).
\end{equation}
Comparing~\ref{equ:spectral_reconstruction_laplacian} and~\ref{equ:spectral_reconstruction_cotangent}, we can find that the term of vertex area can be ignored in our Laplacian2Mesh. 

\section{Quantitative Comparison}

We use the meshes in the Tele-Alien dataset to perform eigendecomposition and spectral reconstruction. There are three metrics we used to evaluate the performances of the two methods, which are the mean of eigendecomposition time (MET), the mean of spectral reconstruction errors (MRE), and the mean of spectral reconstruction time (MRT). 

MET is the average time taken by all meshes in the dataset to perform eigendecomposition, which is defined as
\begin{equation}
\label{equ:average_eigendecomposition_time}
MET(\mathcal{M}_0, \mathcal{M}_0, \cdots, \mathcal{M}_{N_m-1}) = \frac{1}{N_m}\sum_{i=0}^{N_m-1}t_i,
\end{equation}
where $t_i$ is the eigendecomposition time for the mesh $\mathcal{M}_i$, and $N_m$ is the number of mesh models in the dataset.

For spectral reconstruction, the goal is that the mesh reconstructed with fewer eigenvectors has a smaller reconstruction error than the original mesh model, which means that the distance between the vertex coordinates $\widehat{\mathbf{v}}_i$ of the mesh $\widehat{\mathcal{M}}_i$ obtained by spectral reconstruction and the vertex coordinates $\mathbf{v}_i$ of the original mesh $\mathcal{M}_i$ should be as small as possible. Therefore, we perform spectral reconstruction by choosing different numbers of eigenvectors (i.e. $\mathbf{k} = \{751, 400, 100, 80, 40, 20, 10\}$), and then the MRE is 
\begin{equation}
\label{equ:average_reconstruction_error}
MRE(\mathcal{M}_0, \mathcal{M}_0, \cdots, \mathcal{M}_{N_m-1}) = \frac{1}{N_mN_k}\sum_{i=0}^{N_m-1}\sum_{j=0}^{N_k-1}\lVert \widehat{\mathbf{v}}_{ij} - \mathbf{v}_{ij} \rVert,
\end{equation}
where $N_k$ is the number of combinations of eigenvectors, which is $N_k=7$ in our experiment. MRT is the average spectral reconstruction time of the meshes in the dataset with different $k_i$, which is
\begin{equation}
\label{equ:average_reconstruction_time}
MRT(\mathcal{M}_0, \mathcal{M}_0, \cdots, \mathcal{M}_{N_m-1}) = \frac{1}{N_mN_k}\sum_{i=0}^{N_m-1}\sum_{j=0}^{N_k-1}\bar{t}_{ij},
\end{equation}
where $\bar{t}_{ij}$ is the spectral reconstruction time for the mesh $\mathcal{M}_i$ when $k_j$ eigenvectors are chosen.

\renewcommand\arraystretch{1.2}
\begin{table}[htb]
  \caption{The performance of discrete Laplacian matrix and cotangent weight matrix for the task of spectral reconstruction. "MET" is the mean of eigendecomposition time for the meshes in the Tele-Alien dataset, "MRE" is the mean of spectral reconstruction error, and "MRT" is the mean of spectral reconstruction time.}
  \label{tab:reconstruction_error_time}
  \begin{center}
  {
  \begin{tabular}{c
	*{3}{c}%
	}
  \toprule
   \textbf{Method} & \textbf{MET} $\downarrow$ & \textbf{MRE} $\downarrow$ & \textbf{MRT} $\downarrow$ \\
  \midrule
   Laplacian  & 1048ms & $6.55\times10^{-1}$ & 2.18ms \\
   
   cotangent (ours)  & \textbf{824}ms & $\mathbf{6.33\times10^{-1}}$ & \textbf{1.27}ms \\
  \bottomrule
  \end{tabular}}%
  \end{center}
\end{table}

As the results are shown in Table~\ref{tab:reconstruction_error_time}, we can know that we can perform eigendecomposition faster because we do not need to calculate the area term, and our method achieves lower MRE and uses less MRT. The comparison between the eigendecomposition of Laplacian and
the eigendecomposition of cotangent weight matrix experiments validates the effectiveness of our reasoning.

\begin{table}[tb]
\centering
\caption{The mesh segmentation accuracy (Acc.) and the running time on the dense (D.)~\cite{Hu2022SubdivNet} and sparse (S.)~\cite{Wang2012ActiveCO} COSEG-Chairs dataset. The running time (millisecond, ms) includes the preprocessing time (Pre.), training time (Train), and testing time (Test), averaged for each single input of the specified input size.}
\label{tab:chairs-dense}
\begin{tabular}{ccccccc} 
\toprule
\multirow{2}{*}{\textbf{Input Size}} & \multicolumn{2}{c}{\textbf{Chairs}} & \multirow{2}{*}{\textbf{Acc.}} & \multicolumn{3}{c}{\textbf{Times} (ms)}  \\ 
\cmidrule{2-3} 
\cmidrule{5-7}
   & \multicolumn{1}{l}{\textbf{D.}~\cite{Hu2022SubdivNet}} & \multicolumn{1}{l}{\textbf{S.}~\cite{Wang2012ActiveCO}} &  & \textbf{Pre.} & \textbf{Train} & \textbf{Test} \\
\cmidrule{1-7}
$128-64-16$ &  & \checkmark & 90.5$\%$ & 520 & 285 & 243\\
$256-64-16$ &  & \checkmark & 96.6$\%$ & 520 & 391 & 271\\
$256-64-16$ & \checkmark&   & 89.9$\%$ & 5830 & 435 & 281\\ 
$1024-256-64$ & \checkmark& & 95.7$\%$ & 5830 & 887 & 452\\
\bottomrule
\end{tabular}
\end{table}

\section{Dense Meshes}

We compare the mesh segmentation of our Laplacian2Mesh on sparse meshes~\cite{Wang2012ActiveCO} (used in the main paper) and dense meshes~\cite{Hu2022SubdivNet}. Table~\ref{tab:chairs-dense} reports the segmentation results and the running time on sparse meshes in the first two rows, and those on dense meshes in the bottom two rows. The experiment in the last row shows that our method can also achieve gratifying performance on dense meshes. However, as shown in this table, the performance relies on the input size of the network, i.e. the number of eigenvectors used to form our multi-resolution spectral signals. Dense meshes require larger memory space and computation amount to perform the eigenvector decomposition and the transformation between different eigenvector basis. With our current implementation, we can only run the experiment on dense meshes with input size 1024-256-64 in order to prevent the out-of-memory issue. It's worth noting that our method requires at least $N/2$ eigenvectors (input size 4096-1024-256 for the dense mesh dataset) to obtain the best performance, as described and validated in the main paper, which means that our method has the potential to get better results on dense meshes.

The sparse matrix computation techniques are useful to solve this issue. There has been some research~\cite{Nicholas2022DiffusionNet, Smirnov2021HodgeNetLS} using sparse computation for eigenvector decomposition and spectral signal processing. Implementing our Laplacian2Mesh with the sparse matrix computation techniques will further improve the performance.

\ifCLASSOPTIONcaptionsoff
  \newpage
\fi

\bibliographystyle{IEEEtran}
\bibliography{references.bib}

\begin{thebibliography}{100}
\providecommand{\url}[1]{#1}
\csname url@samestyle\endcsname
\providecommand{\newblock}{\relax}
\providecommand{\bibinfo}[2]{#2}
\providecommand{\BIBentrySTDinterwordspacing}{\spaceskip=0pt\relax}
\providecommand{\BIBentryALTinterwordstretchfactor}{4}
\providecommand{\BIBentryALTinterwordspacing}{\spaceskip=\fontdimen2\font plus
\BIBentryALTinterwordstretchfactor\fontdimen3\font minus
  \fontdimen4\font\relax}
\providecommand{\BIBforeignlanguage}[2]{{%
\expandafter\ifx\csname l@#1\endcsname\relax
\typeout{** WARNING: IEEEtran.bst: No hyphenation pattern has been}%
\typeout{** loaded for the language `#1'. Using the pattern for}%
\typeout{** the default language instead.}%
\else
\language=\csname l@#1\endcsname
\fi
#2}}
\providecommand{\BIBdecl}{\relax}
\BIBdecl

\bibitem{Qi2017PointNetDL}
C.~Qi, H.~Su, K.~Mo, and L.~J. Guibas, ``Pointnet: Deep learning on point sets
  for 3d classification and segmentation,'' \emph{2017 IEEE Conference on
  Computer Vision and Pattern Recognition (CVPR)}, pp. 77--85, 2017.

\bibitem{Qi2017PointNetDH}
C.~Qi, L.~Yi, H.~Su, and L.~J. Guibas, ``Pointnet++: Deep hierarchical feature
  learning on point sets in a metric space,'' in \emph{NIPS}, 2017.

\bibitem{Li2018PointCNNCO}
Y.~Li, R.~Bu, M.~Sun, W.~Wu, X.~Di, and B.~Chen, ``Pointcnn: Convolution on
  x-transformed points,'' in \emph{NeurIPS}, 2018.

\bibitem{Su2015MultiviewCN}
H.~Su, S.~Maji, E.~Kalogerakis, and E.~G. Learned-Miller, ``Multi-view
  convolutional neural networks for 3d shape recognition,'' \emph{2015 IEEE
  International Conference on Computer Vision (ICCV)}, pp. 945--953, 2015.

\bibitem{Le2017AMR}
T.~Le, G.~Bui, and Y.~Duan, ``A multi-view recurrent neural network for 3d mesh
  segmentation,'' \emph{Comput. Graph.}, vol.~66, pp. 103--112, 2017.

\bibitem{Klokov2017EscapeFC}
R.~Klokov and V.~S. Lempitsky, ``Escape from cells: Deep kd-networks for the
  recognition of 3d point cloud models,'' \emph{2017 IEEE International
  Conference on Computer Vision (ICCV)}, pp. 863--872, 2017.

\bibitem{Wang2017OCNN}
P.-S. Wang, Y.~Liu, Y.-X. Guo, C.-Y. Sun, and X.~Tong, ``O-cnn,'' \emph{ACM
  Transactions on Graphics (TOG)}, vol.~36, pp. 1 -- 11, 2017.

\bibitem{Hanocka2019MeshCNNAN}
R.~Hanocka, A.~Hertz, N.~Fish, R.~Giryes, S.~Fleishman, and D.~Cohen-Or,
  ``Meshcnn: a network with an edge,'' \emph{ACM Transactions on Graphics
  (TOG)}, vol.~38, pp. 1 -- 12, 2019.

\bibitem{Lahav2020MeshWalkerDM}
A.~Lahav and A.~Tal, ``Meshwalker: Deep mesh understanding by random walks,''
  \emph{ACM Trans. Graph.}, vol.~39, pp. 263:1--263:13, 2020.

\bibitem{Karpathy2014LargeScaleVC}
A.~Karpathy, G.~Toderici, S.~Shetty, T.~Leung, R.~Sukthankar, and L.~Fei-Fei,
  ``Large-scale video classification with convolutional neural networks,''
  \emph{2014 IEEE Conference on Computer Vision and Pattern Recognition}, pp.
  1725--1732, 2014.

\bibitem{Chen2017ReferenceBL}
M.~Chen, G.~Ding, S.~Zhao, H.~Chen, Q.~Liu, and J.~Han, ``Reference based lstm
  for image captioning,'' in \emph{AAAI}, 2017.

\bibitem{Shelhamer2017FullyCN}
E.~Shelhamer, J.~Long, and T.~Darrell, ``Fully convolutional networks for
  semantic segmentation,'' \emph{IEEE Transactions on Pattern Analysis and
  Machine Intelligence}, vol.~39, pp. 640--651, 2017.

\bibitem{Wu20153DSA}
Z.~Wu, S.~Song, A.~Khosla, F.~Yu, L.~Zhang, X.~Tang, and J.~Xiao, ``3d
  shapenets: A deep representation for volumetric shapes,'' \emph{2015 IEEE
  Conference on Computer Vision and Pattern Recognition (CVPR)}, pp.
  1912--1920, 2015.

\bibitem{Botsch2010PolygonMP}
M.~Botsch, L.~P. Kobbelt, M.~Pauly, P.~Alliez, and B.~L{\'e}vy, ``Polygon mesh
  processing,'' 2010.

\bibitem{Le2018PointGridAD}
T.~Le and Y.~Duan, ``Pointgrid: A deep network for 3d shape understanding,''
  \emph{2018 IEEE/CVF Conference on Computer Vision and Pattern Recognition},
  pp. 9204--9214, 2018.

\bibitem{Hu2021VMNetVN}
Z.~Hu, X.~Bai, J.~Shang, R.~Zhang, J.~Dong, X.~Wang, G.~Sun, H.~Fu, and C.-L.
  Tai, ``Vmnet: Voxel-mesh network for geodesic-aware 3d semantic
  segmentation,'' \emph{ArXiv}, vol. abs/2107.13824, 2021.

\bibitem{Wang2020VoxSegNetVC}
Z.~Wang and F.~Lu, ``Voxsegnet: Volumetric cnns for semantic part segmentation
  of 3d shapes,'' \emph{IEEE Transactions on Visualization and Computer
  Graphics}, vol.~26, pp. 2919--2930, 2020.

\bibitem{Maturana2015VoxNetA3}
D.~Maturana and S.~A. Scherer, ``Voxnet: A 3d convolutional neural network for
  real-time object recognition,'' \emph{2015 IEEE/RSJ International Conference
  on Intelligent Robots and Systems (IROS)}, pp. 922--928, 2015.

\bibitem{Redmon2016YouOL}
J.~Redmon, S.~K. Divvala, R.~B. Girshick, and A.~Farhadi, ``You only look once:
  Unified, real-time object detection,'' \emph{2016 IEEE Conference on Computer
  Vision and Pattern Recognition (CVPR)}, pp. 779--788, 2016.

\bibitem{Dosovitskiy2021AnII}
A.~Dosovitskiy, L.~Beyer, A.~Kolesnikov, D.~Weissenborn, X.~Zhai,
  T.~Unterthiner, M.~Dehghani, M.~Minderer, G.~Heigold, S.~Gelly, J.~Uszkoreit,
  and N.~Houlsby, ``An image is worth 16x16 words: Transformers for image
  recognition at scale,'' \emph{ArXiv}, vol. abs/2010.11929, 2021.

\bibitem{Lvy2006LaplaceBeltramiET}
B.~L{\'e}vy, ``Laplace-beltrami eigenfunctions towards an algorithm that
  "understands" geometry,'' \emph{IEEE International Conference on Shape
  Modeling and Applications 2006 (SMI'06)}, pp. 13--13, 2006.

\bibitem{He2020CurvaNetGD}
W.~He, Z.~Jiang, C.~Zhang, and A.~M. Sainju, ``Curvanet: Geometric deep
  learning based on directional curvature for 3d shape analysis,''
  \emph{Proceedings of the 26th ACM SIGKDD International Conference on
  Knowledge Discovery \& Data Mining}, 2020.

\bibitem{Qiao2022LearningO3}
Y.-L. Qiao, L.~Gao, J.~Yang, P.~L. Rosin, Y.-K. Lai, and X.~Chen, ``Learning on
  3d meshes with laplacian encoding and pooling,'' \emph{IEEE Transactions on
  Visualization and Computer Graphics}, vol.~28, pp. 1317--1327, 2022.

\bibitem{Hu2020SqueezeandExcitationN}
J.~Hu, L.~Shen, S.~Albanie, G.~Sun, and E.~Wu, ``Squeeze-and-excitation
  networks,'' \emph{IEEE Transactions on Pattern Analysis and Machine
  Intelligence}, vol.~42, pp. 2011--2023, 2020.

\bibitem{Bronstein2021GeometricDL}
M.~M. Bronstein, J.~Bruna, T.~Cohen, and P.~Velivckovi'c, ``Geometric deep
  learning: Grids, groups, graphs, geodesics, and gauges,'' \emph{ArXiv}, vol.
  abs/2104.13478, 2021.

\bibitem{Bronstein2017GeometricDL}
M.~M. Bronstein, J.~Bruna, Y.~LeCun, A.~D. Szlam, and P.~Vandergheynst,
  ``Geometric deep learning: Going beyond euclidean data,'' \emph{IEEE Signal
  Processing Magazine}, vol.~34, pp. 18--42, 2017.

\bibitem{Xiao2020ASO}
Y.~Xiao, Y.-K. Lai, F.-L. Zhang, C.~Li, and L.~Gao, ``A survey on deep geometry
  learning: From a representation perspective,'' \emph{Computational Visual
  Media}, vol.~6, pp. 113--133, 2020.

\bibitem{Brock2016GenerativeAD}
A.~Brock, T.~Lim, J.~M. Ritchie, and N.~Weston, ``Generative and discriminative
  voxel modeling with convolutional neural networks,'' \emph{ArXiv}, vol.
  abs/1608.04236, 2016.

\bibitem{Tchapmi2017SEGCloudSS}
L.~P. Tchapmi, C.~B. Choy, I.~Armeni, J.~Gwak, and S.~Savarese, ``Segcloud:
  Semantic segmentation of 3d point clouds,'' \emph{2017 International
  Conference on 3D Vision (3DV)}, pp. 537--547, 2017.

\bibitem{Hanocka2019ALIGNetPA}
R.~Hanocka, N.~Fish, Z.~Wang, R.~Giryes, S.~Fleishman, and D.~Cohen-Or,
  ``Alignet: Partial-shape agnostic alignment via unsupervised learning,''
  \emph{ACM Trans. Graph.}, vol.~38, pp. 1:1--1:14, 2019.

\bibitem{Li2016FPNNFP}
Y.~Li, S.~Pirk, H.~Su, C.~Qi, and L.~J. Guibas, ``Fpnn: Field probing neural
  networks for 3d data,'' \emph{ArXiv}, vol. abs/1605.06240, 2016.

\bibitem{Riegler2017OctNetLD}
G.~Riegler, A.~O. Ulusoy, and A.~Geiger, ``Octnet: Learning deep 3d
  representations at high resolutions,'' \emph{2017 IEEE Conference on Computer
  Vision and Pattern Recognition (CVPR)}, pp. 6620--6629, 2017.

\bibitem{Graham20183DSS}
B.~Graham, M.~Engelcke, and L.~van~der Maaten, ``3d semantic segmentation with
  submanifold sparse convolutional networks,'' \emph{2018 IEEE/CVF Conference
  on Computer Vision and Pattern Recognition}, pp. 9224--9232, 2018.

\bibitem{Sinha2016DeepL3}
A.~Sinha, J.~Bai, and K.~Ramani, ``Deep learning 3d shape surfaces using
  geometry images,'' in \emph{ECCV}, 2016.

\bibitem{Maron2017ConvolutionalNN}
H.~Maron, M.~Galun, N.~Aigerman, M.~Trope, N.~Dym, E.~Yumer, V.~G. Kim, and
  Y.~Lipman, ``Convolutional neural networks on surfaces via seamless toric
  covers,'' \emph{ACM Transactions on Graphics (TOG)}, vol.~36, pp. 1 -- 10,
  2017.

\bibitem{Haim2019SurfaceNV}
N.~Haim, N.~Segol, H.~Ben-Hamu, H.~Maron, and Y.~Lipman, ``Surface networks via
  general covers,'' \emph{2019 IEEE/CVF International Conference on Computer
  Vision (ICCV)}, pp. 632--641, 2019.

\bibitem{Sfikas2017ExploitingTP}
K.~Sfikas, T.~Theoharis, and I.~Pratikakis, ``Exploiting the panorama
  representation for convolutional neural network classification and
  retrieval,'' in \emph{3DOR@Eurographics}, 2017.

\bibitem{Shi2015DeepPanoDP}
B.~Shi, S.~Bai, Z.~Zhou, and X.~Bai, ``Deeppano: Deep panoramic representation
  for 3-d shape recognition,'' \emph{IEEE Signal Processing Letters}, vol.~22,
  pp. 2339--2343, 2015.

\bibitem{Kalogerakis20173DSS}
E.~Kalogerakis, M.~Averkiou, S.~Maji, and S.~Chaudhuri, ``3d shape segmentation
  with projective convolutional networks,'' \emph{2017 IEEE Conference on
  Computer Vision and Pattern Recognition (CVPR)}, pp. 6630--6639, 2017.

\bibitem{Masci2015GeodesicCN}
J.~Masci, D.~Boscaini, M.~M. Bronstein, and P.~Vandergheynst, ``Geodesic
  convolutional neural networks on riemannian manifolds,'' \emph{2015 IEEE
  International Conference on Computer Vision Workshop (ICCVW)}, pp. 832--840,
  2015.

\bibitem{Monti2017GeometricDL}
F.~Monti, D.~Boscaini, J.~Masci, E.~Rodol{\`a}, J.~Svoboda, and M.~M.
  Bronstein, ``Geometric deep learning on graphs and manifolds using mixture
  model cnns,'' \emph{2017 IEEE Conference on Computer Vision and Pattern
  Recognition (CVPR)}, pp. 5425--5434, 2017.

\bibitem{Verma2018FeaStNetFG}
N.~Verma, E.~Boyer, and J.~Verbeek, ``Feastnet: Feature-steered graph
  convolutions for 3d shape analysis,'' \emph{2018 IEEE/CVF Conference on
  Computer Vision and Pattern Recognition}, pp. 2598--2606, 2018.

\bibitem{Boscaini2015LearningCD}
D.~Boscaini, J.~Masci, S.~Melzi, M.~M. Bronstein, U.~Castellani, and
  P.~Vandergheynst, ``Learning class‐specific descriptors for deformable
  shapes using localized spectral convolutional networks,'' \emph{Computer
  Graphics Forum}, vol.~34, 2015.

\bibitem{Fey2018SplineCNNFG}
M.~Fey, J.~E. Lenssen, F.~Weichert, and H.~M{\"u}ller, ``Splinecnn: Fast
  geometric deep learning with continuous b-spline kernels,'' \emph{2018
  IEEE/CVF Conference on Computer Vision and Pattern Recognition}, pp.
  869--877, 2018.

\bibitem{Schonsheck2018ParallelTC}
S.~C. Schonsheck, B.~Dong, and R.~Lai, ``Parallel transport convolution: A new
  tool for convolutional neural networks on manifolds,'' \emph{ArXiv}, vol.
  abs/1805.07857, 2018.

\bibitem{Schult2020DualConvMeshNetJG}
J.~Schult, F.~Engelmann, T.~Kontogianni, and B.~Leibe, ``Dualconvmesh-net:
  Joint geodesic and euclidean convolutions on 3d meshes,'' \emph{2020 IEEE/CVF
  Conference on Computer Vision and Pattern Recognition (CVPR)}, pp.
  8609--8619, 2020.

\bibitem{Kipf2017SemiSupervisedCW}
T.~Kipf and M.~Welling, ``Semi-supervised classification with graph
  convolutional networks,'' \emph{ArXiv}, vol. abs/1609.02907, 2017.

\bibitem{Wang2019DynamicGC}
Y.~Wang, Y.~Sun, Z.~Liu, S.~E. Sarma, M.~M. Bronstein, and J.~M. Solomon,
  ``Dynamic graph cnn for learning on point clouds,'' \emph{ACM Transactions on
  Graphics (TOG)}, vol.~38, pp. 1 -- 12, 2019.

\bibitem{Atwood2016DiffusionConvolutionalNN}
J.~Atwood and D.~F. Towsley, ``Diffusion-convolutional neural networks,'' in
  \emph{NIPS}, 2016.

\bibitem{Milano2020PrimalDualMC}
F.~Milano, A.~Loquercio, A.~Rosinol, D.~Scaramuzza, and L.~Carlone,
  ``Primal-dual mesh convolutional neural networks,'' \emph{ArXiv}, vol.
  abs/2010.12455, 2020.

\bibitem{Hu2022SubdivNet}
\BIBentryALTinterwordspacing
S.-M. Hu, Z.-N. Liu, M.-H. Guo, J.-X. Cai, J.~Huang, T.-J. Mu, and R.~R.
  Martin, ``Subdivision-based mesh convolution networks,'' \emph{ACM Trans.
  Graph.}, vol.~41, no.~3, mar 2022. [Online]. Available:
  \url{https://doi.org/10.1145/3506694}
\BIBentrySTDinterwordspacing

\bibitem{Ovsjanikov2008GlobalIS}
M.~Ovsjanikov, J.~Sun, and L.~J. Guibas, ``Global intrinsic symmetries of
  shapes,'' \emph{Computer Graphics Forum}, vol.~27, 2008.

\bibitem{FunctionalMaps2012}
\BIBentryALTinterwordspacing
M.~Ovsjanikov, M.~Ben-Chen, J.~Solomon, A.~Butscher, and L.~Guibas,
  ``Functional maps: A flexible representation of maps between shapes,''
  \emph{ACM Trans. Graph.}, vol.~31, no.~4, jul 2012. [Online]. Available:
  \url{https://doi.org/10.1145/2185520.2185526}
\BIBentrySTDinterwordspacing

\bibitem{Bronstein2011ShapeRW}
M.~M. Bronstein and A.~M. Bronstein, ``Shape recognition with spectral
  distances,'' \emph{IEEE Transactions on Pattern Analysis and Machine
  Intelligence}, vol.~33, pp. 1065--1071, 2011.

\bibitem{Reuter2006LaplaceBeltramiSA}
M.~Reuter, F.-E. Wolter, and N.~Peinecke, ``Laplace-beltrami spectra as
  'shape-dna' of surfaces and solids,'' \emph{Comput. Aided Des.}, vol.~38, pp.
  342--366, 2006.

\bibitem{Bronstein2011ShapeGG}
A.~M. Bronstein, M.~M. Bronstein, L.~J. Guibas, and M.~Ovsjanikov, ``Shape
  google: Geometric words and expressions for invariant shape retrieval,''
  \emph{ACM Trans. Graph.}, vol.~30, pp. 1:1--1:20, 2011.

\bibitem{Sun2009ACA}
J.~Sun, M.~Ovsjanikov, and L.~J. Guibas, ``A concise and provably informative
  multi‐scale signature based on heat diffusion,'' \emph{Computer Graphics
  Forum}, vol.~28, 2009.

\bibitem{Wang2019IntrinsicAE}
Y.~Wang and J.~M. Solomon, ``Intrinsic and extrinsic operators for shape
  analysis,'' \emph{Handbook of Numerical Analysis}, 2019.

\bibitem{Liu2007MeshSV}
R.~Liu and H.~Zhang, ``Mesh segmentation via spectral embedding and contour
  analysis,'' \emph{Computer Graphics Forum}, vol.~26, 2007.

\bibitem{Bruna2014SpectralNA}
J.~Bruna, W.~Zaremba, A.~D. Szlam, and Y.~LeCun, ``Spectral networks and
  locally connected networks on graphs,'' \emph{CoRR}, vol. abs/1312.6203,
  2014.

\bibitem{Pinkall1993ComputingDM}
U.~Pinkall and K.~Polthier, ``Computing discrete minimal surfaces and their
  conjugates,'' \emph{Exp. Math.}, vol.~2, pp. 15--36, 1993.

\bibitem{10.2312:SGP:SGP03:127-137}
X.~Gu and S.-T. Yau, ``{Global Conformal Surface Parameterization},'' in
  \emph{Eurographics Symposium on Geometry Processing}, L.~Kobbelt,
  P.~Schroeder, and H.~Hoppe, Eds.\hskip 1em plus 0.5em minus 0.4em\relax The
  Eurographics Association, 2003.

\bibitem{Botsch2008OnLV}
M.~Botsch and O.~Sorkine-Hornung, ``On linear variational surface deformation
  methods,'' \emph{IEEE Transactions on Visualization and Computer Graphics},
  vol.~14, pp. 213--230, 2008.

\bibitem{He2016DeepRL}
K.~He, X.~Zhang, S.~Ren, and J.~Sun, ``Deep residual learning for image
  recognition,'' \emph{2016 IEEE Conference on Computer Vision and Pattern
  Recognition (CVPR)}, pp. 770--778, 2016.

\bibitem{Clevert2016FastAA}
D.-A. Clevert, T.~Unterthiner, and S.~Hochreiter, ``Fast and accurate deep
  network learning by exponential linear units (elus),'' \emph{arXiv:
  Learning}, 2016.

\bibitem{Ronneberger2015UNetCN}
O.~Ronneberger, P.~Fischer, and T.~Brox, ``U-net: Convolutional networks for
  biomedical image segmentation,'' in \emph{MICCAI}, 2015.

\bibitem{Hoppe1997ViewdependentRO}
H.~Hoppe, ``View-dependent refinement of progressive meshes,''
  \emph{Proceedings of the 24th annual conference on Computer graphics and
  interactive techniques}, 1997.

\bibitem{Boykov2001InteractiveGC}
Y.~Boykov and M.~Jolly, ``Interactive graph cuts for optimal boundary \& region
  segmentation of objects in n-d images,'' \emph{Proceedings Eighth IEEE
  International Conference on Computer Vision. ICCV 2001}, vol.~1, pp. 105--112
  vol.1, 2001.

\bibitem{Lian2011SHRECT}
Z.~Lian, A.~Godil, B.~Bustos, M.~Daoudi, J.~Hermans, S.~Kawamura, Y.~Kurita,
  G.~Lavou{\'e}, H.~V. Nguyen, R.~Ohbuchi, Y.~Ohkita, Y.~Ohishi, F.~M. Porikli,
  M.~Reuter, I.~Sipiran, D.~Smeets, P.~Suetens, H.~Tabia, and D.~Vandermeulen,
  ``Shrec '11 track: Shape retrieval on non-rigid 3d watertight meshes,'' in
  \emph{3DOR@Eurographics}, 2011.

\bibitem{Ezuz2017GWCNNAM}
D.~Ezuz, J.~M. Solomon, V.~G. Kim, and M.~Ben-Chen, ``Gwcnn: A metric alignment
  layer for deep shape analysis,'' \emph{Computer Graphics Forum}, vol.~36,
  2017.

\bibitem{Wang2012ActiveCO}
Y.~Wang, S.~Asafi, O.~M. van Kaick, H.~Zhang, D.~Cohen-Or, and B.~Chen,
  ``Active co-analysis of a set of shapes,'' \emph{ACM Transactions on Graphics
  (TOG)}, vol.~31, pp. 1 -- 10, 2012.

\bibitem{Kalogerakis2010Learning3M}
E.~Kalogerakis, A.~Hertzmann, and K.~Singh, ``Learning 3d mesh segmentation and
  labeling,'' in \emph{SIGGRAPH 2010}, 2010.

\bibitem{Anguelov2005SCAPESC}
D.~Anguelov, P.~Srinivasan, D.~Koller, S.~Thrun, J.~Rodgers, and J.~Davis,
  ``Scape: shape completion and animation of people,'' \emph{ACM Trans.
  Graph.}, vol.~24, pp. 408--416, 2005.

\bibitem{Bogo2014FAUSTDA}
F.~Bogo, J.~Romero, M.~Loper, and M.~J. Black, ``Faust: Dataset and evaluation
  for 3d mesh registration,'' \emph{2014 IEEE Conference on Computer Vision and
  Pattern Recognition}, pp. 3794--3801, 2014.

\bibitem{Vlasic2008ArticulatedMA}
D.~Vlasic, I.~Baran, W.~Matusik, and J.~Popovi{\'c}, ``Articulated mesh
  animation from multi-view silhouettes,'' \emph{ACM SIGGRAPH 2008 papers},
  2008.

\bibitem{Adobe16}
Adobe, ``Adobe fuse 3d characters,'' https://www.mixamo.com, 2016.

\bibitem{giorgi2007shape}
D.~Giorgi, S.~Biasotti, and L.~Paraboschi, ``Shape retrieval contest 2007:
  Watertight models track,'' \emph{SHREC competition}, vol.~8, no.~7, p.~7,
  2007.

\bibitem{Smirnov2021HodgeNetLS}
D.~Smirnov and J.~M. Solomon, ``Hodgenet: Learning spectral geometry on
  triangle meshes,'' \emph{ACM Trans. Graph.}, vol.~40, pp. 166:1--166:11,
  2021.

\bibitem{Latecki2000ShapeSM}
L.~J. Latecki and R.~Lak{\"a}mper, ``Shape similarity measure based on
  correspondence of visual parts,'' \emph{IEEE Trans. Pattern Anal. Mach.
  Intell.}, vol.~22, pp. 1185--1190, 2000.

\bibitem{Liu2020DISTRD}
S.~Liu, Y.~Zhang, S.~Peng, B.~Shi, M.~Pollefeys, and Z.~Cui, ``Dist: Rendering
  deep implicit signed distance function with differentiable sphere tracing,''
  \emph{2020 IEEE/CVF Conference on Computer Vision and Pattern Recognition
  (CVPR)}, pp. 2016--2025, 2020.

\bibitem{Guo2021PCTPC}
M.-H. Guo, J.~Cai, Z.-N. Liu, T.-J. Mu, R.~R. Martin, and S.~Hu, ``Pct: Point
  cloud transformer,'' \emph{Comput. Vis. Media}, vol.~7, pp. 187--199, 2021.

\bibitem{feng2019meshnet}
Y.~Feng, Y.~Feng, H.~You, X.~Zhao, and Y.~Gao, ``Meshnet: Mesh neural network
  for 3d shape representation,'' in \emph{Proceedings of the AAAI Conference on
  Artificial Intelligence}, vol.~33, 2019, pp. 8279--8286.

\bibitem{Xu2017DCN}
H.~Xu, M.~Dong, and Z.~Zhong, ``Directionally convolutional networks for 3d
  shape segmentation,'' in \emph{2017 IEEE International Conference on Computer
  Vision (ICCV)}, 2017, pp. 2717--2726.

\bibitem{Nicholas2022DiffusionNet}
\BIBentryALTinterwordspacing
N.~Sharp, S.~Attaiki, K.~Crane, and M.~Ovsjanikov, ``Diffusionnet:
  Discretization agnostic learning on surfaces,'' \emph{ACM Trans. Graph.},
  vol.~41, no.~3, mar 2022. [Online]. Available:
  \url{https://doi.org/10.1145/3507905}
\BIBentrySTDinterwordspacing

\bibitem{Lehoucq1998ARPACKUG}
R.~B. Lehoucq, D.~C. Sorensen, and C.~Yang, ``Arpack users' guide - solution of
  large-scale eigenvalue problems with implicitly restarted arnoldi methods,''
  in \emph{Software, environments, tools}, 1998.

\bibitem{Rustamov2007LaplaceBeltramiEF}
R.~M. Rustamov, ``Laplace-beltrami eigenfunctions for deformation invariant
  shape representation,'' in \emph{Symposium on Geometry Processing}, 2007.

\bibitem{Liu2004SegmentationO3}
R.~Liu and H.~Zhang, ``Segmentation of 3d meshes through spectral clustering,''
  \emph{12th Pacific Conference on Computer Graphics and Applications, 2004. PG
  2004. Proceedings.}, pp. 298--305, 2004.

\bibitem{Katz2003HierarchicalMD}
S.~Katz and A.~Tal, ``Hierarchical mesh decomposition using fuzzy clustering
  and cuts,'' \emph{ACM SIGGRAPH 2003 Papers}, 2003.

\bibitem{Shi1997NormalizedCA}
J.~Shi and J.~Malik, ``Normalized cuts and image segmentation,''
  \emph{Proceedings of IEEE Computer Society Conference on Computer Vision and
  Pattern Recognition}, pp. 731--737, 1997.

\bibitem{Hamilton2017InductiveRL}
W.~L. Hamilton, Z.~Ying, and J.~Leskovec, ``Inductive representation learning
  on large graphs,'' in \emph{NIPS}, 2017.

\bibitem{Zhou2022LetsAD}
Q.~Zhou, M.~Li, Q.~Zeng, A.~Aristidou, X.~Zhang, L.~Chen, and C.~Tu, ``Let’s
  all dance: Enhancing amateur dance motions,'' 2022.

\bibitem{Thomas2019KPConvFA}
H.~Thomas, C.~Qi, J.-E. Deschaud, B.~Marcotegui, F.~Goulette, and L.~J. Guibas,
  ``Kpconv: Flexible and deformable convolution for point clouds,'' \emph{2019
  IEEE/CVF International Conference on Computer Vision (ICCV)}, pp. 6410--6419,
  2019.

\bibitem{estrach2014spectral}
J.~B. Estrach, W.~Zaremba, A.~Szlam, and Y.~LeCun, ``Spectral networks and deep
  locally connected networks on graphs,'' in \emph{2nd International conference
  on learning representations, ICLR}, vol. 2014, 2014.

\bibitem{defferrard2016convolutional}
M.~Defferrard, X.~Bresson, and P.~Vandergheynst, ``Convolutional neural
  networks on graphs with fast localized spectral filtering,'' \emph{Advances
  in neural information processing systems}, vol.~29, 2016.

\bibitem{Liu2021DeepIM}
S.~Liu, H.~Guo, H.~Pan, P.-S. Wang, X.~Tong, and Y.~Liu, ``Deep implicit moving
  least-squares functions for 3d reconstruction,'' \emph{2021 IEEE/CVF
  Conference on Computer Vision and Pattern Recognition (CVPR)}, pp.
  1788--1797, 2021.

\bibitem{Meyer2002DiscreteDO}
M.~Meyer, M.~Desbrun, P.~Schr{\"o}der, and A.~H. Barr, ``Discrete
  differential-geometry operators for triangulated 2-manifolds,'' in
  \emph{VisMath}, 2002.

\bibitem{Sharp2020ALF}
N.~Sharp and K.~Crane, ``A laplacian for nonmanifold triangle meshes,''
  \emph{Computer Graphics Forum}, vol.~39, 2020.

\bibitem{Wang-2017-ocnn}
P.-S. Wang, C.-Y. Sun, Y.~Liu, and X.~Tong, ``{Adaptive O-CNN: A Patch-based
  Deep Representation of 3D Shapes},'' \emph{ACM Transactions on Graphics
  (SIGGRAPH Asia)}, vol.~37, no.~6, 2018.

\bibitem{Lvy2010SpectralMP}
B.~L{\'e}vy and H.~Zhang, ``Spectral mesh processing,'' in \emph{SIGGRAPH '10},
  2010.

\bibitem{Wardetzky2007DiscreteLO}
M.~Wardetzky, S.~Mathur, F.~K{\"a}lberer, and E.~Grinspun, ``Discrete laplace
  operators: no free lunch,'' in \emph{Symposium on Geometry Processing}, 2007.

\bibitem{Hanocka2021AnIT}
R.~Hanocka and H.-T.~D. Liu, ``An introduction to deep learning on meshes,''
  \emph{ACM SIGGRAPH 2021 Courses}, 2021.

\bibitem{Loop1987SmoothSS}
C.~T. Loop, ``Smooth subdivision surfaces based on triangles,'' in
  \emph{Masters Thesis, Department of Mathematics, University of Utah (January
  1987)}, 1987.

\bibitem{Litman2014LearningSD}
R.~Litman and A.~M. Bronstein, ``Learning spectral descriptors for deformable
  shape correspondence,'' \emph{IEEE Transactions on Pattern Analysis and
  Machine Intelligence}, vol.~36, pp. 171--180, 2014.

\bibitem{Karni2000SpectralCO}
Z.~Karni and C.~Gotsman, ``Spectral compression of mesh geometry,''
  \emph{Proceedings of the 27th annual conference on Computer graphics and
  interactive techniques}, 2000.

\bibitem{He2018BagOT}
T.~He, Z.~Zhang, H.~Zhang, Z.~Zhang, J.~Xie, and M.~Li, ``Bag of tricks for
  image classification with convolutional neural networks,'' \emph{2019
  IEEE/CVF Conference on Computer Vision and Pattern Recognition (CVPR)}, pp.
  558--567, 2018.

\end{thebibliography}

\end{document}